\def\year{2021}\relax
\documentclass[letterpaper]{article} 
\usepackage{aaai21}  
\usepackage{times}  
\usepackage{helvet} 
\usepackage{courier}  
\usepackage[hyphens]{url}  
\usepackage{graphicx} 
\usepackage{algorithm}
\usepackage{algorithmic}
\usepackage{amssymb}
\usepackage{amsmath}
\usepackage[backref]{hyperref} 

\usepackage{epsfig}
\usepackage{graphicx}
\usepackage{multirow}
\usepackage{tabularx}
\usepackage{booktabs}
\usepackage{url}
\usepackage{makecell}

\usepackage{float}
\usepackage{subfigure}
\usepackage{graphicx}

\usepackage{natbib}  
\usepackage{caption} 
\title{PP-OCR: A Practical Ultra Lightweight OCR System}
\author {
    Yuning Du, Chenxia Li, Ruoyu Guo, Xiaoting Yin, Weiwei Liu,   \\
    Jun Zhou, Yifan Bai, Zilin Yu, Yehua Yang, Qingqing Dang, Haoshuang Wang \\
}
\affiliations {
    Baidu Inc. \\
    \{duyuning, yangyehua\}@baidu.com
}

\begin{document}

\maketitle

\begin{abstract}
The Optical Character Recognition (OCR) systems have been widely used in various of application scenarios, such as office automation (OA) systems, factory automations, online educations, map productions etc. However, OCR is still a challenging task due to the various of text appearances and the demand of computational efficiency. In this paper, we propose a practical ultra lightweight OCR system, i.e., PP-OCR. The overall model size of the PP-OCR is only 3.5M for recognizing 6622 Chinese characters and 2.8M for recognizing 63 alphanumeric symbols, respectively. We introduce a bag of strategies to either enhance the model ability or reduce the model size. The corresponding ablation experiments with the real data are also provided. Meanwhile, several pre-trained models for the Chinese and English recognition are released, including a text detector (97K images are used), a direction classifier (600K images are used) as well as a text recognizer (17.9M images are used). Besides, the proposed PP-OCR are also verified in several other language recognition tasks, including French, Korean, Japanese and German. All of the above mentioned models are open-sourced and the codes are available in the GitHub repository, i.e., https://github.com/PaddlePaddle/PaddleOCR.
\end{abstract}

\section{Introduction}

\begin{figure}[t]
\centering
\subfigure{
\begin{minipage}[t]{0.85\linewidth}
\centering
\includegraphics[width=\columnwidth]{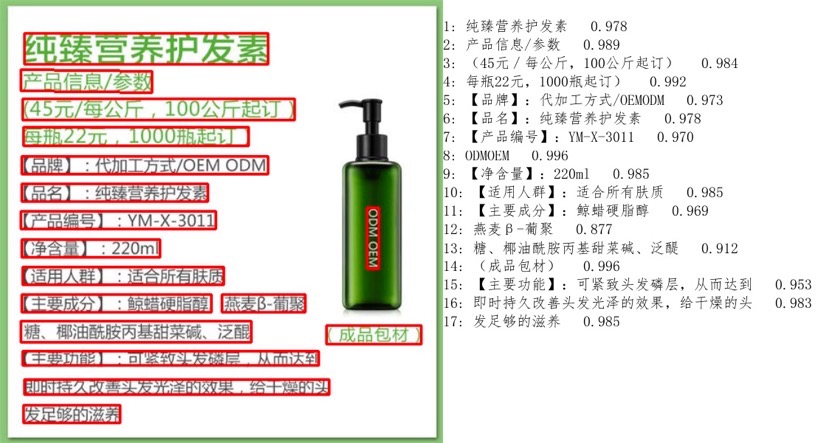}
\end{minipage}
}

\subfigure{
\begin{minipage}[t]{0.85\linewidth}
\centering
\includegraphics[width=\columnwidth]{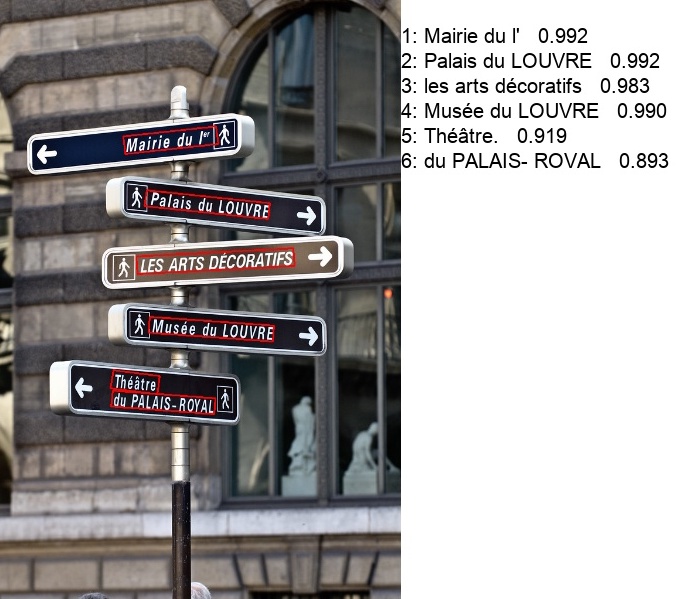}
\end{minipage}
}

\caption{Some image results of the proposed PP-OCR system.}
\label{res_p1}
\end{figure}

OCR (Optical Character Recognition), a technology which targets at recognizing text in images automatically as shown in Figure \ref{res_p1}, has a long research history and a wide range of application scenarios, such as document electronization, identity authentication, digital financial system, and vehicle license plate recognition. Moreover, in factory, products can be more conveniently managed by extracting the text information of products automatically. Students’ offline homework or test paper can be electronized with an OCR system to make the communication between teachers and students more efficient. OCR can also be used for labeling the point of interests (POI) of a street view image,  benefiting the map production efficiency. Rich application scenarios endow OCR technology with great commercial value, meanwhile, a lot of challenges.

\begin{figure*}[t]
\centering
\includegraphics[width=15cm]{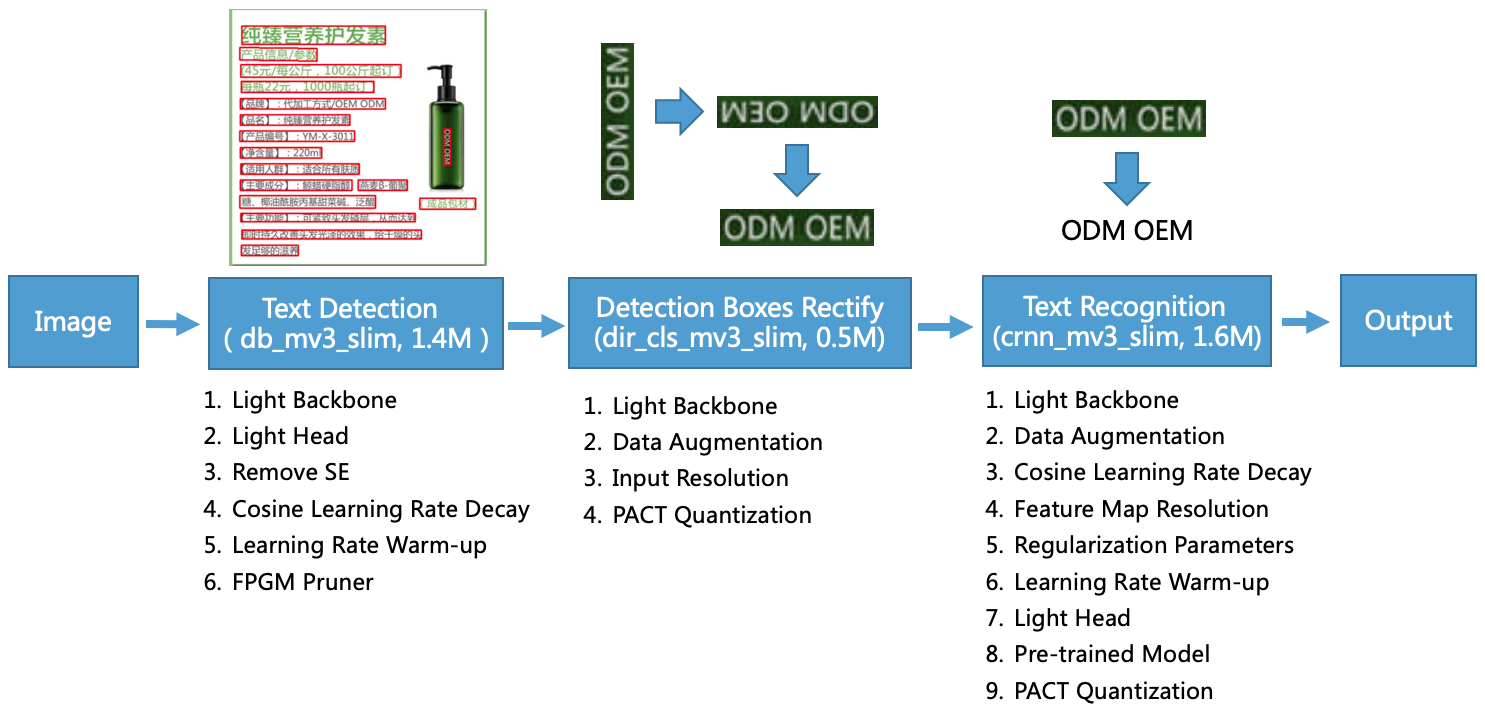}
\caption{The framework of the proposed PP-OCR. The model size in the figure is about Chinese and English characters recognition. For alphanumeric symbols recognition, the model size of text recognition is from 1.6M to 0.9M. The rest of the models are the same size.}
\label{framework}
\end{figure*}

\begin{figure}[t]
\centering
\includegraphics[width=\columnwidth]{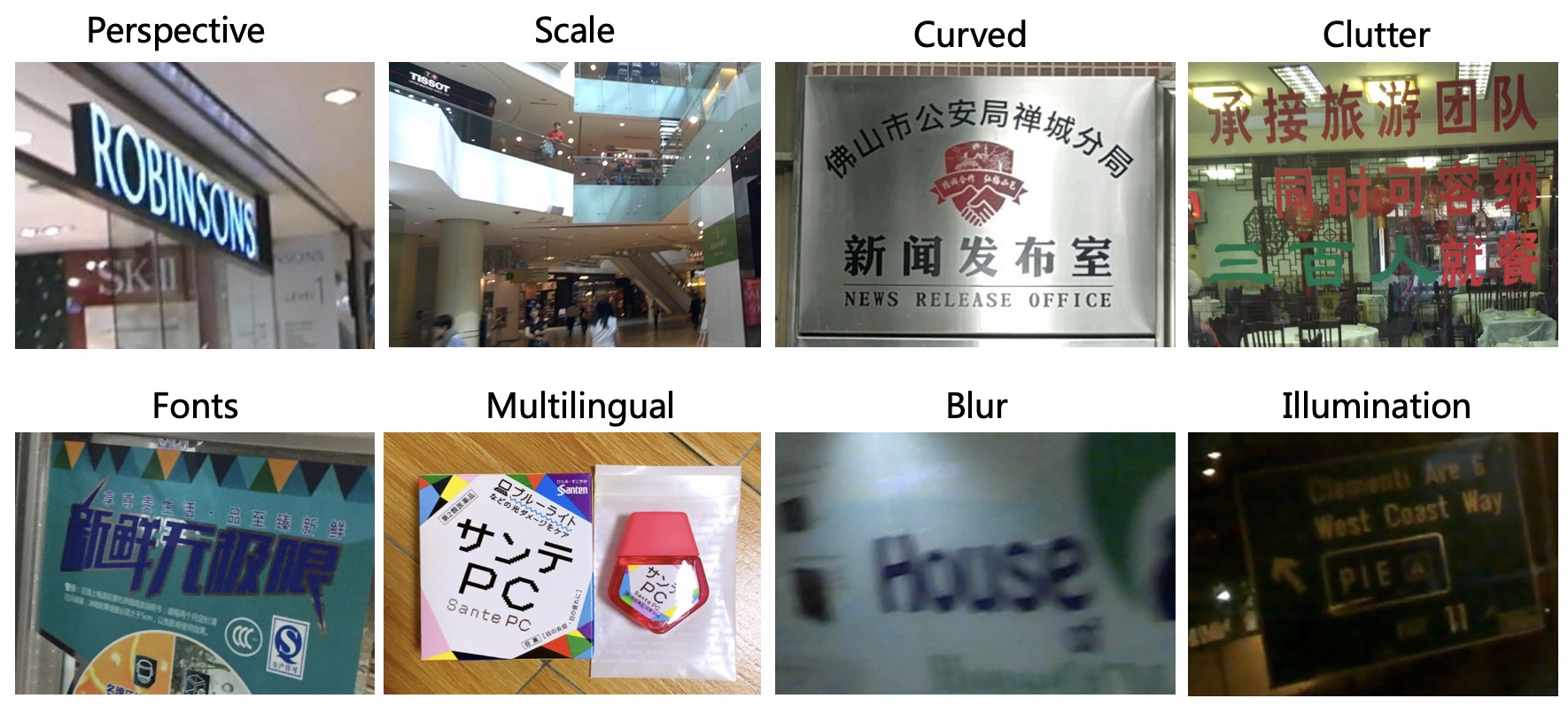}
\caption{Some images contained scene text.}
\label{ex_SRT}
\end{figure}

\begin{figure}[t]
\centering
\includegraphics[width=\columnwidth]{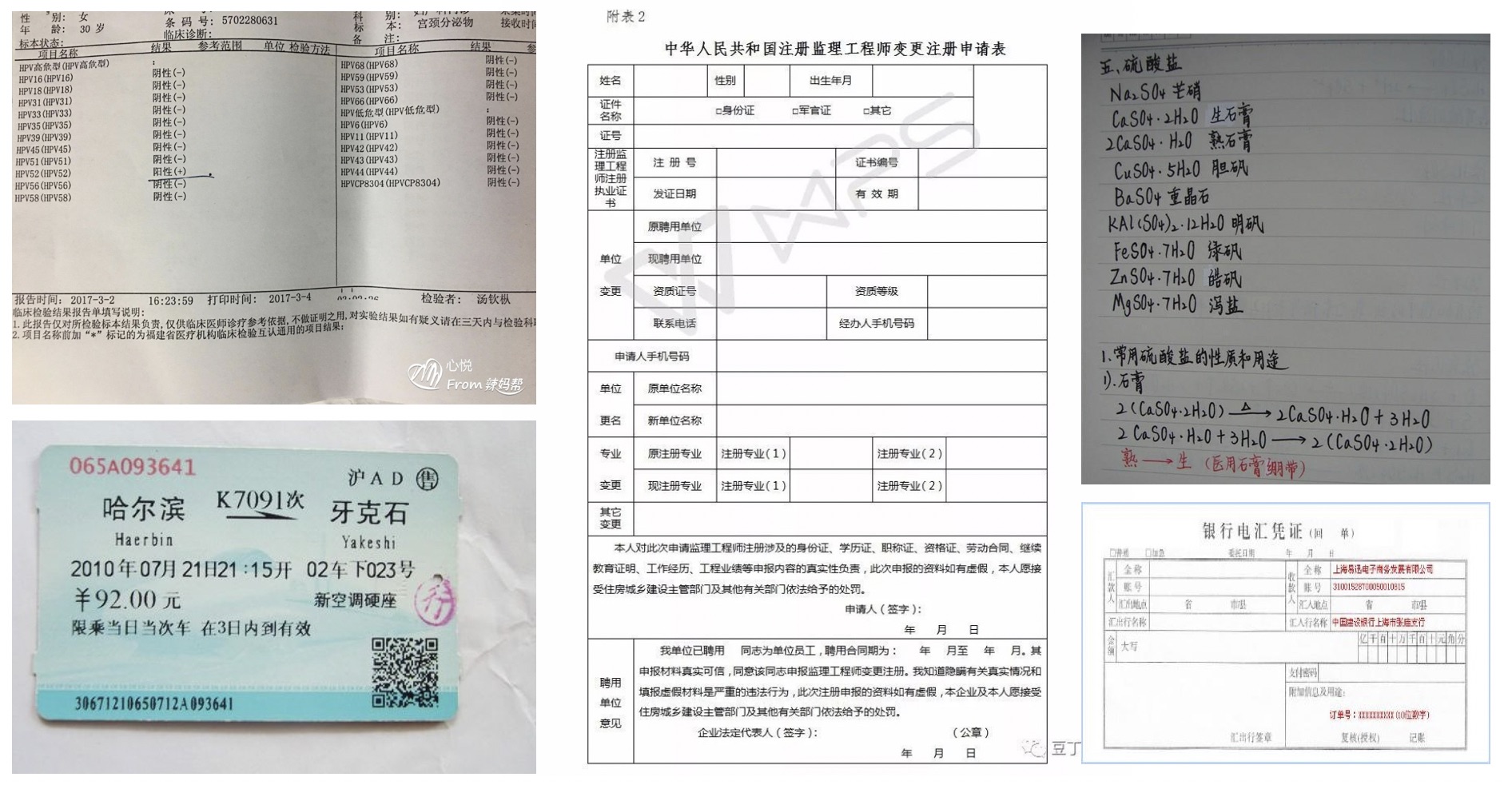}
\caption{Some images contained document text.}
\label{ex_doc}
\end{figure}

\begin{figure*}[t]
\centering
\includegraphics[width=14.5cm]{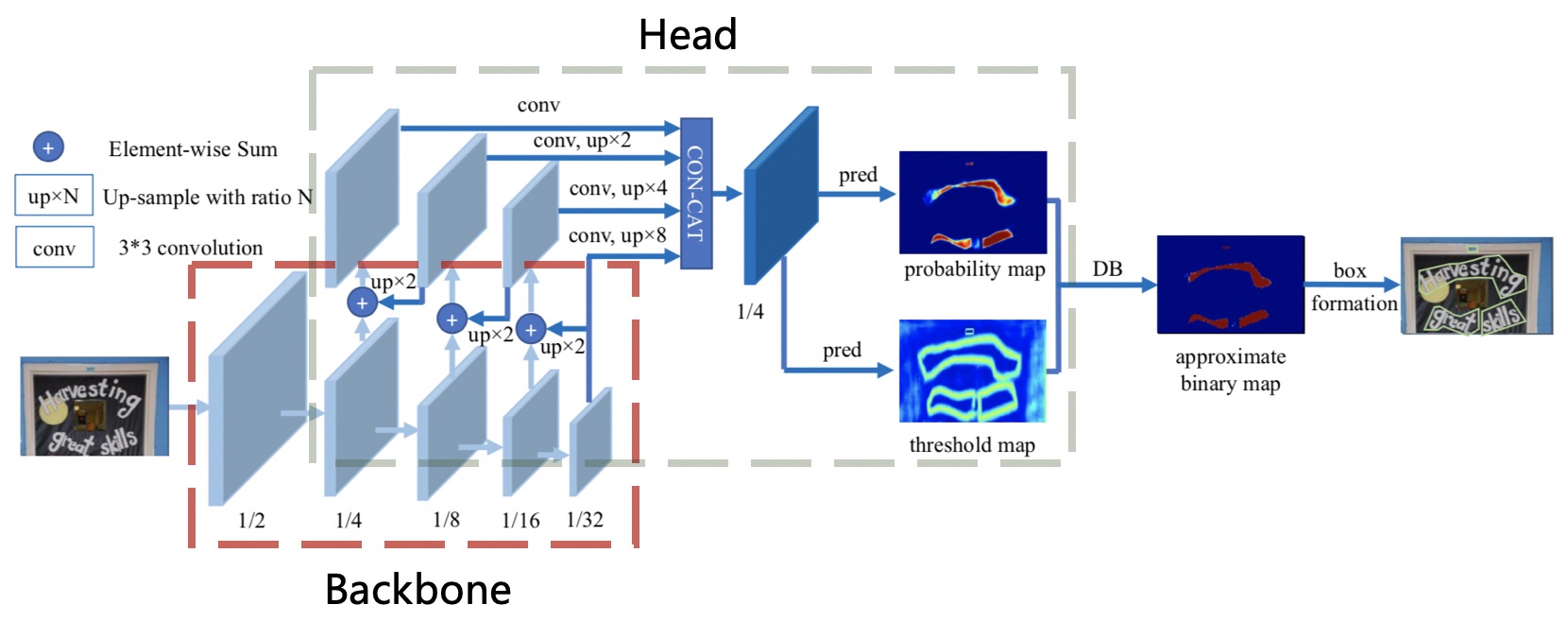}
\caption{Architecture of the text detector DB. This figure comes from the paper of DB \cite{liao2020real}. The red and gray rectangles show the backbone and head of the text detector separately.}
\label{db_framework}
\end{figure*}

\textbf{Various of Text Appearances} Text in image can be generally divided into two categories: scene text and document text. Scene text refers to the text in natural scene as shown in Figure \ref{ex_SRT}, which usually changes dramatically for the factors such as perspective, scaling, bending, clutter, fonts, multilingual, blur, illumination, etc. Document text, as shown in Figure \ref{ex_doc}, is more often encountered in practical application. Different problems caused by the high density and long text need to be solved. Otherwise, document image text recognition often comes with the need to structure the results, which introduced a new hard task.

\textbf{Computational Efficiency} In practical, the images that need to be processed are usually massive, which makes high computational efficiency an important criterion for designing an OCR system. CPU is preferred to be used than GPU considering the cost. In particular, the OCR system need to be run on embedded devices in many scenarios, such as cell phones, which makes it necessary to consider the model size. Trade off model size and performance is difficult but of great value. In this paper, we propose a practical ultra lightweight OCR system, named as PP-OCR, which consists of three parts, text detection, detected boxes rectification and text recognition as shown in Figure \ref{framework}.

\textbf{Text Detection} The purpose of text detection is to locate the text area in the image. In PP-OCR, we use Differentiable Binarization (DB) \cite{liao2020real} as text detector which is based on a simple segmentation network. The simple post-processing of DB makes it very efficient. In order to further improve its effectiveness and efficiency, the following six strategies are used: light backbone, light head, remove SE module, cosine learning rate decay, learning rate warm-up, and FPGM pruner. Finally, the model size of the text detector is reduced to 1.4M.

\textbf{Detection Boxes Rectify} Before recognizing the detected text, the text box needs to be transformed into a horizontal rectangle box for subsequent text recognition, which is easy to be achieved by geometric transformation as the detection frame is composed of four points. However, the rectified boxes may be reversed. Thus, a classifier is needed to determine the text direction. If a box is determined reversed, further flipping is required. Training a text direction classifier is a simple image classification task. We adopt the following four strategies to enhance the model ability and reduce the model size: light backbone, data augmentation, input resolution and PACT quantization. Finally, the model size of the text direction classifier is 500KB.

\textbf{Text Recognition} In PP-OCR, we use CRNN \cite{shi2016end} as text recognizer, which is widely used and practical for text recognition. CRNN integrates feature extraction and sequence modeling. It adopts the Connectionist Temporal Classification(CTC) loss to avoid the inconsistency between prediction and label. To enhance the model ability and reduce the model size of a text recognizer, the following nine strategies are used: light backbone, data augmentation, cosine learning rate decay, feature map resolution, regularization parameters, learning rate warm-up, light head, pre-trained model and PACT quantization. Finally, the model size of the text recognizer is only 1.6M for Chinese and English recognition and 900KB for alphanumeric symbols recognition.

In order to implement a practical OCR system, we construct a large-scale dataset for Chinese and English recognition as an example. Specifically, text detection dataset has 97K images. Direction classification dataset has 600k images. Text recognition dataset has 17.9M images. A small amount of the data are selected to conduct ablation experiments quickly and choose the appropriate strategies. We make a lot of ablation experiments to show the effects of different strategies in Figure \ref{framework}. Besides, we also verify the proposed PP-OCR system for other languages recognition which including alphanumeric symbols, French, Korean, Japanese and German.

The rest of the paper is organized as follows. In section 2, we present the bag of model enhancement or slimming strategies. Experimental results are discussed in section 3 and conclusion is conducted in section 4.

\section{Enhancement or Slimming Strategies}

\subsection{Text Detection}
In this section, the details of six strategies for enhancing the model ability or reducing the model size of a text detector will be introduced. Figure \ref{db_framework} shows the architecture of the text detector DB.

\begin{figure}[t]
\centering
\includegraphics[width=\columnwidth]{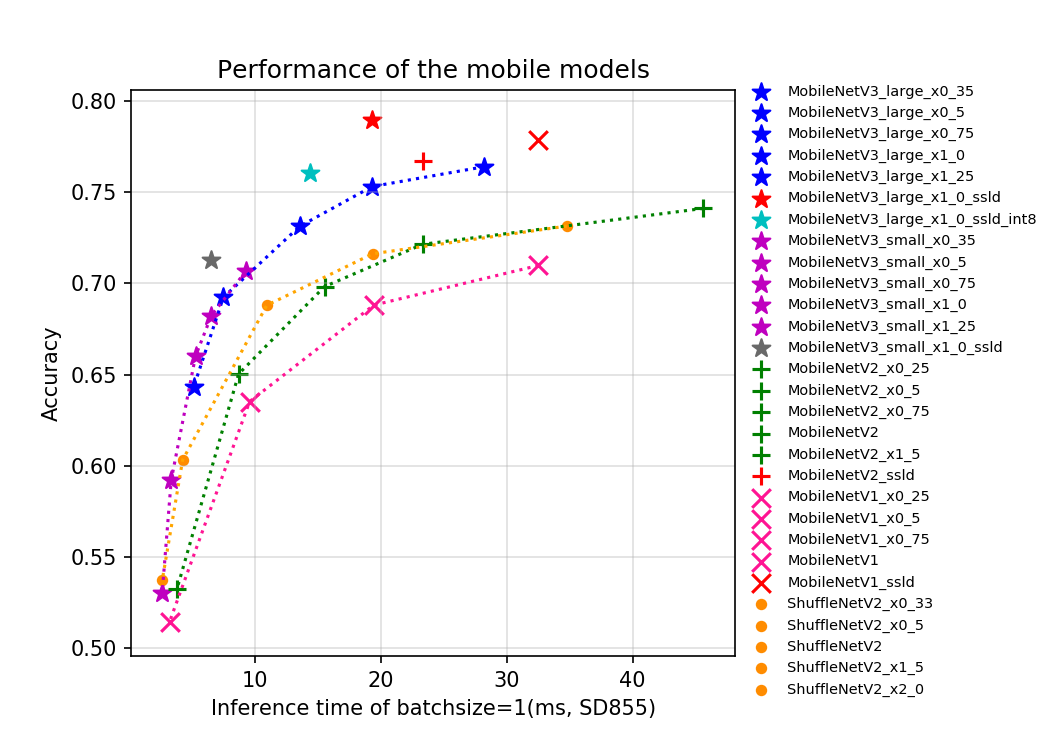}
\caption{The performance of some light backbones on the ImageNet 1000 classification, including MobileNetV1, MobileNetV2, MobileNetV3 and ShuffleNetV2 series. The inference time is tested on Snapdragon 855 (SD855) with the batch size set as 1.}
\label{cls_mobile_acc}
\end{figure}

\textbf{Light Backbone} The size of backbone has dominant effect on the model size of a text detector. Therefore, light backbones should be selected for building the ultra lightweight models. With the development of image classification, MobileNetV1, MobileNetV2, MobileNetV3 and ShuffleNetV2 series are often used as the light backbones. Each series has different scale. Thanks to the inference time on CPU and accuracy of more than 20 kinds of backbones are provided by PaddleClas\footnote[1]{https://github.com/PaddlePaddle/PaddleClas/}, as shown in Figure \ref{cls_mobile_acc}, MobileNetV3 can achieve higher accuracy when the predict time are same. As for the choice of scale, we adopt MobileNetV3\_large\_x0.5 to balance accuracy and efficiency empirically. Incidentally, PaddleClas provides a total of up to 24 series of image classification network structures and training configurations, 122 models' pretrained weights and their evaluation metrics, such as ResNet, ResNet\_vd, SEResNeXt, Res2Net, Res2Net\_vd, DPN, DenseNet, EfficientNet, Xception, HRNet, etc.  

\begin{figure}[t]
\centering
\includegraphics[width=\columnwidth]{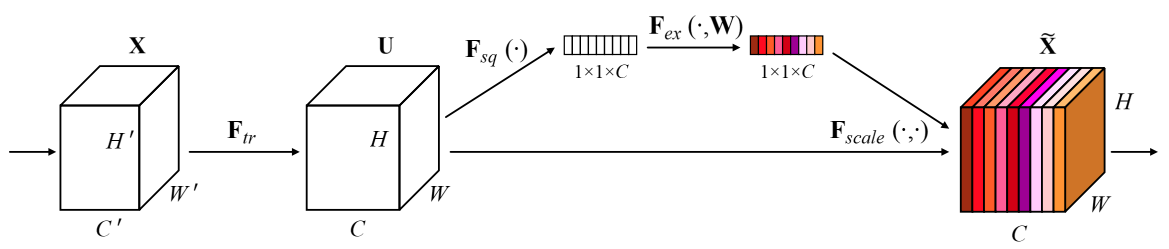}
\caption{Architecture of the SE block. This figure comes from the paper \cite{hu2018squeeze}.}
\label{se_arc}
\end{figure}

\begin{figure}[t]
\centering
\includegraphics[width=\columnwidth]{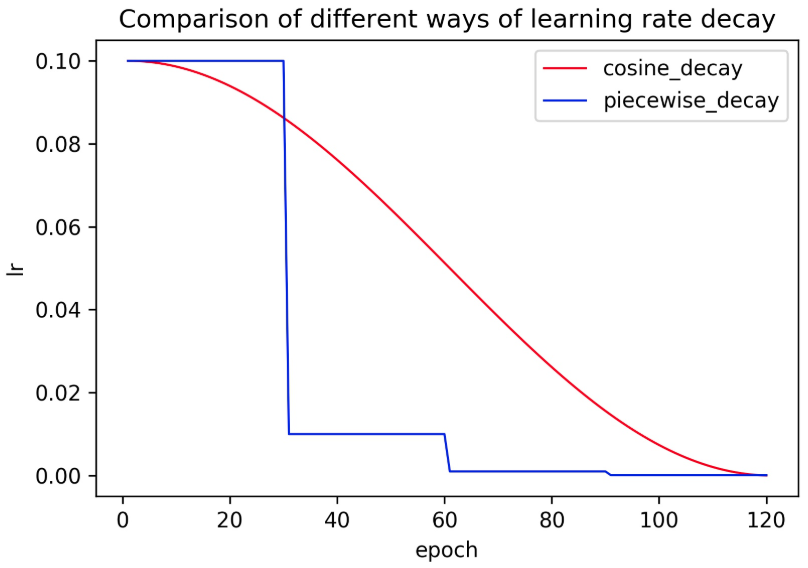}
\caption{Comparison of different ways of learning rate decay.}
\label{cos}
\end{figure}

\begin{figure}[t]
\centering
\includegraphics[width=\columnwidth]{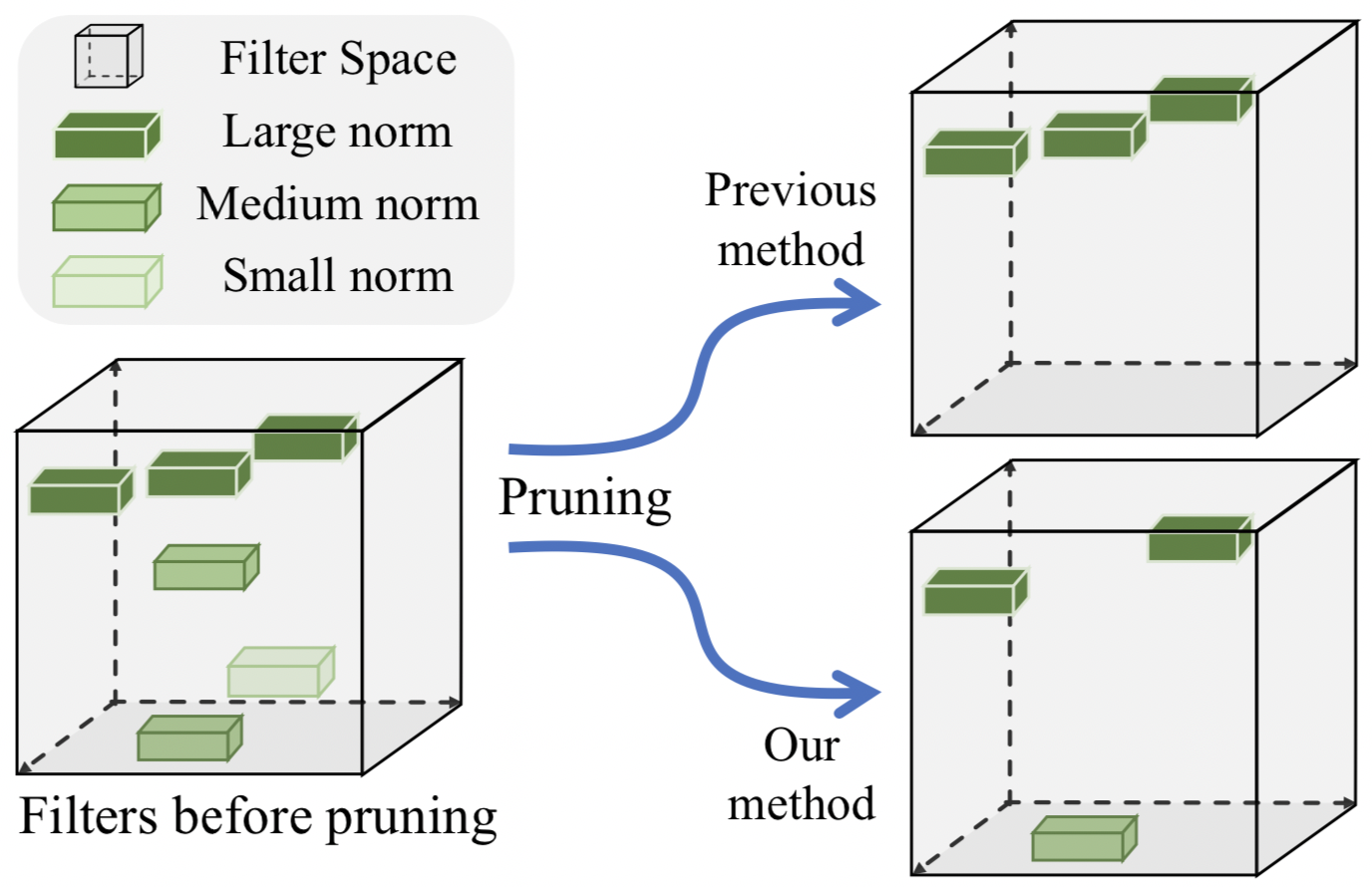}
\caption{Illustration of FPGM Pruner. This figure comes from the paper \cite{he2019filter}.}
\label{det_pruning}
\end{figure}

\textbf{Light Head} The head of the text detector is similar as the FPN \cite{lin2017feature} architecture in object detection and fuse the feature maps of the different scales to improve the effect for the small text regions detection. For convenience of merging the different resolution feature maps, $1\times1$ convolution is often used to reduce the feature maps to the same number of channel (we use inner\_channels for short). The probability map and the threshold map are generated from the fused feature map with convolutions which are also associated with the above inner\_channels. Thus inner\_channels has a great influence on the model size. When inner\_channels is reduced from 256 to 96, the model size is reduced from 7M to 4.1M, but the accuracy declines slightly.

\textbf{Remove SE} SE is the short for squeeze-and-excitation \cite{hu2018squeeze}. As shown in Figure \ref{se_arc}, SE blocks model inter-dependencies between channels explicitly and re-calibrate channel-wise feature responses adaptively. Because SE blocks can improve the accuracy of the vision tasks obviously, the search space of MobileNetV3 contains them and numerous of SE blocks are in MobileNetV3 architecture. However, when the input resolution is large, such as $640\times640$, it is hard to estimate the channel-wise feature responses with the SE block. The accuracy improvement is limited, but the time cost is very high. When the SE blocks are removed from the backbone, the model size is reduced from 4.1M to 2.5M, but the accuracy has no effect.

\textbf{Cosine Learning Rate Decay} The learning rate is the hyperparameter to control the learning speed. The lower the learning rate, the slower the change of the loss value. Though using a low learning rate can ensure that you will not miss any local minimum, but it also means that the convergence speed is slow. In the early stage of training, the weights are in random initialization state, so we can set a relatively large learning rate for faster convergence. In the late stage of training, the weights are close to the optimal values, so a relatively smaller learning rate should be used. Cosine learning rate decay has become the preferred learning rate reduction strategy for improving model accuracy. During the entire training process, cosine learning rate decay keeps a relatively large learning rate, so its convergence is slower, but the final convergence accuracy is better. Figure \ref{cos} compares the different ways of learning rate decay.

\textbf{Learning Rate Warm-up} The paper \cite{he2019bag} shows that using learning rate warm-up operation can help to improve the accuracy in the image classification. At the beginning of the training process, using a too large learning rate may result in numerical instability, a small learning rate is recommended to be used. When the training process is stable, the initial learning rate is to be used. For text detection, the experiments show that this strategy also is effective.

\textbf{FPGM Pruner} Pruning is another method to improve the inference efficiency of neural network model. In order to avoid the model performance degradation caused by the model pruning, we use FPGM \cite{he2019filter} to find the unimportant sub-network in original models. FPGM uses geometric median as the criterion and the each filter in a convolution layer is considered as a point in Euclidean space. Then calculate the geometric median of these points and remove the filters with the similar values, as shown in Figure \ref{det_pruning}. The compress ratio of each layer is also important for pruning a model. Pruning every layer uniformly usually leads to significant performance degradation. In PP-OCR, the pruning sensitivity of each layer is calculated according to the method in \cite{li2016pruning} and then used to evaluate the redundancy of each layer.

\subsection{Direction Classification}
In this section, the details of four strategies for enhancing the model ability or reducing the model size of a direction classifier will be introduced.

\textbf{Light Backbone} We also adopt MobileNetV3 as the backbone of the direction classifier which is the same as the text detector. Because this task is relatively simple, we use MobileNetV3\_small\_x0.35 to balance accuracy and efficiency empirically. When using larger backbones, the accuracy doesn't improve more.

\textbf{Data Augmentation} This paper \cite{yu2020towards} shows some image processing operations to train a text recognizer, such as rotation, perspective distortion, motion blur and Gaussian noise. Those processes are referred to as BDA (Base Data Augmentation) for short. They are randomly added to the training images. The experiment shows that BDA also is useful for the direction classifier training. Besides BDA, some new data augmentation operations are proposed recently for improving the effect of image classification, for example, AutoAugment \cite{cubuk2019autoaugment}, RandAugment \cite{cubuk2020randaugment}, CutOut \cite{devries2017improved}, RandErasing \cite{zhong2020random}, HideAndSeek \cite{singh2017hide}, GridMask \cite{chen2020gridmask}, Mixup \cite{zhang2017mixup} and Cutmix \cite{yun2019cutmix}. But the experiments show that most of them don't work for the direction classifier training except for RandAugment and RandErasing. RandAugment works best. Eventually, we add BDA and RandAugment to the training images of the direction classification. 

\textbf{Input Resolution} In general, when the input resolution of a normalized image is increased, accuracy will also be improved. Since the backbone of the direction classifier is very light, increasing the resolution properly will not lead to the computation time raise obviously. In the most of the previous text recognition methods, the height and width of a normalized image is set as $32$ and $100$, respectively. However, in PP-OCR, the height and width is set as $48$ and $192$, respectively, to improve the accuracy of the direction classifier. 

\textbf{PACT Quantization} Quantization allows the neural network model to have lower latency, smaller volume and lower computational power consumption. At present, quantization is mainly divided into two categories: offline quantization and online quantization. Offline quantization refers to a fixed-point quantization method that uses methods such as KL divergence and moving average to determine quantization parameters and does not require retraining. Online quantization is to determine quantization parameters during the training process, which can provide less quantization loss than offline quantization mode.

PACT (PArameterized Clipping acTivation) \cite{choi2018pact} is a new online quantification method that removes some outliers from the activations in advance. After removing the outliers, the model can learn more appropriate quantitative scales. The formula for PACT to preprocess the activations is as follows:

\begin{small}
\begin{equation}
y=PACT(x)=0.5(|x|-|x-\alpha |+\alpha )=\left\{\begin{matrix}0\quad x\in  (-\infty, 0) 
\\x \quad x\in  [0, \alpha)
\\\alpha  \quad x\in  [\alpha, +\infty )
\end{matrix}\right.
\end{equation}
\end{small}

The preprocessing of the activation value of the ordinary PACT method is based on the ReLU function. All activation values greater than a certain threshold are truncated. However, the activation functions in MobileNetV3 are not only ReLU, but also hard swish. Using ordinary PACT quantization leads to a higher quantization loss. Therefore, we modify the formula of the activations preprocessing as follows to reduce the quantization loss.

\begin{small}
\begin{equation}
y=PACT(x)=\left\{\begin{matrix}-\alpha \quad x\in  (-\infty,-\alpha) 
\\x \quad x\in  [-\alpha, \alpha)
\\\alpha  \quad x\in  [\alpha, +\infty )
\end{matrix}\right.
\end{equation}
\end{small}

We used the improved PACT quantification method to quantify the direction classifier model. In addition, L2 regularization with a coefficient of 0.001 is added to the PACT parameters to improve the model robustness.

The implementation of the above FPGM Pruner and PACT quantization is based on PaddleSlim\footnote[1]{https://github.com/PaddlePaddle/PaddleSlim/}. PaddleSlim is a toolkit for model compression. It contains a collection of compression strategies, such as pruning, fixed point quantization, knowledge distillation, hyperparameter searching neural architecture search.

\begin{figure}[t]
\centering
\includegraphics[width=\columnwidth]{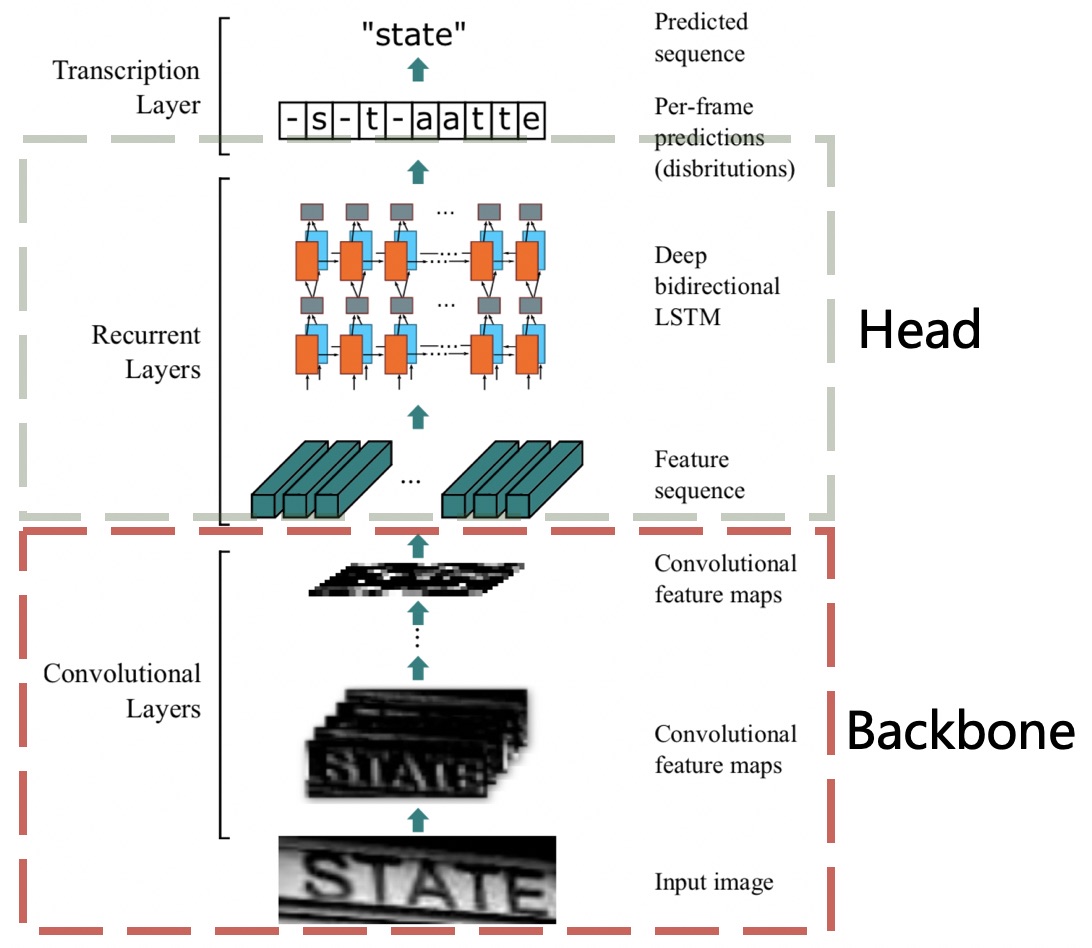}
\caption{Architecture of the text recognizer CRNN. This figure comes from the paper \cite{shi2016end}. The red and gray rectangles show the backbone and head of the text recognizer separately.}
\label{crnn}
\end{figure}

\subsection{Text Recognition}
In this section, the details of nine strategies for enhancing the model ability or reducing the model size of a text recognizer will be introduced. Figure \ref{crnn} shows the architecture of the text recognizer CRNN.

\textbf{Light Backbone} We also adopt MobileNetV3 as the backbone of the text recognizer which is the same as the text detection. MobileNetV3\_small\_x0.5 is selected to balance accuracy and efficiency empirically. If you're not that sensitive to the model size, MobileNetV3\_small\_x1.0 is also a good choice. The model size is just increased by 2M, the accuracy is improved obviously.

\begin{figure}[t]
\centering
\includegraphics[width=\columnwidth]{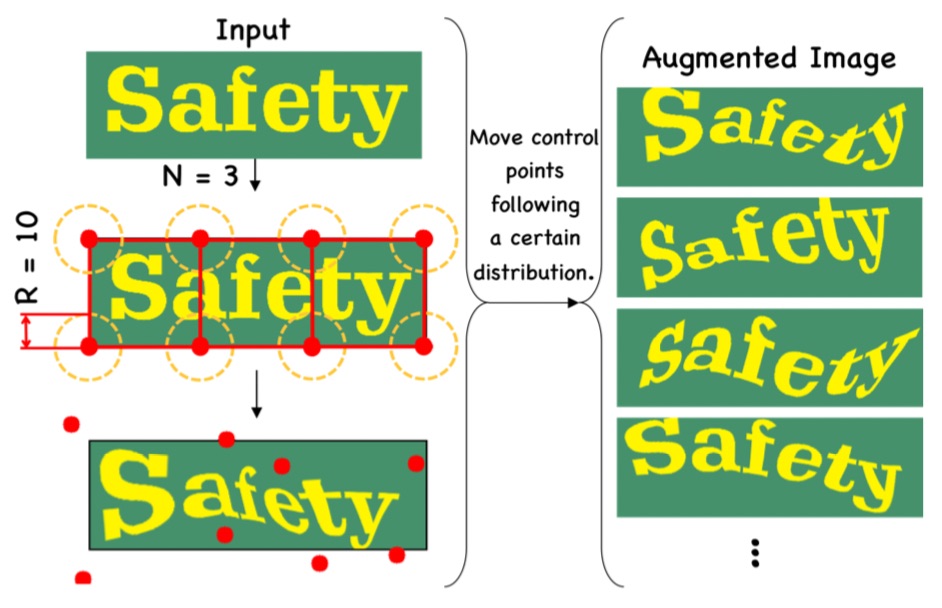}
\caption{Illustration of data augmentation, TIA. This figure comes from the paper \cite{luo2020learn}.}
\label{tia}
\end{figure}

\textbf{Data Augmentation} Besides BDA (Base Data Augmentation) which is often used in text recognition as mentioned earlier, TIA \cite{luo2020learn} also is an effective data augmentation method for text recognition. As shown in Figure \ref{tia}, at first, a set of fiducial points are initialized on the image. Then move the points randomly to generate a new image with the geometric transformation. In PP-OCR, we add BDA and TIA to the training images of the text recognition.

\begin{figure}[t]
\centering
\includegraphics[width=0.9\columnwidth]{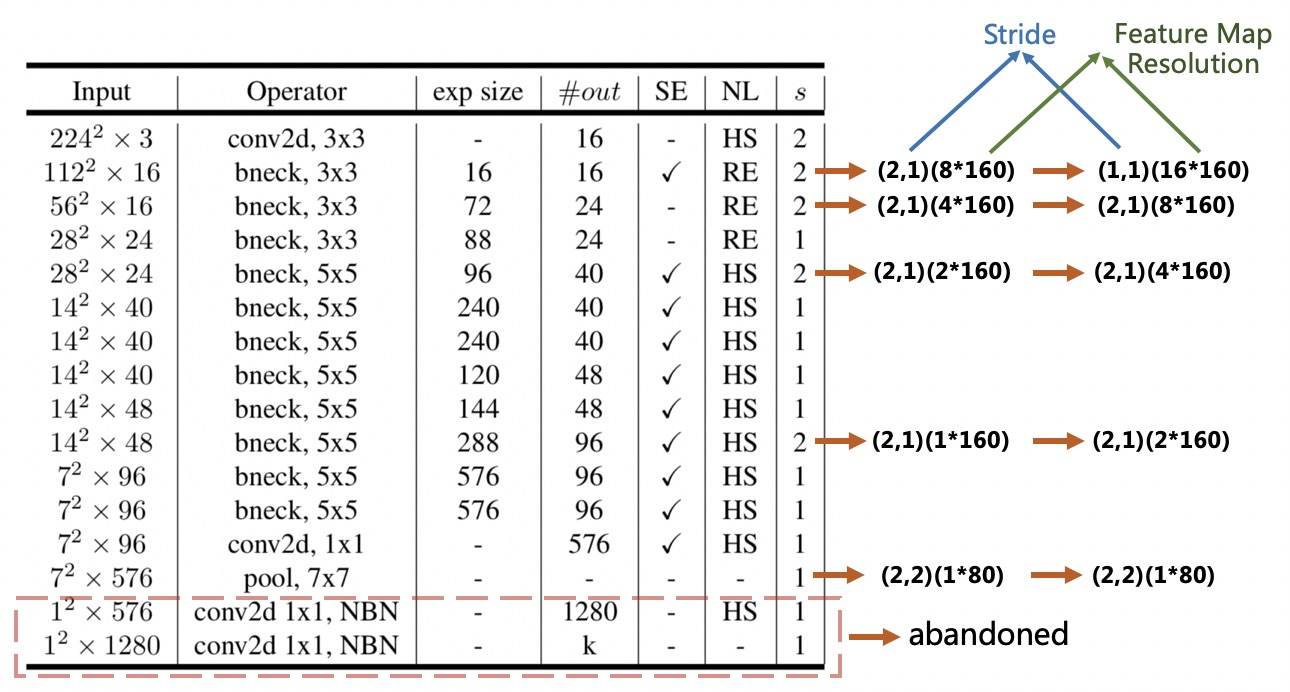}
\caption{Illustration of the modify of the feature map resolution. The table comes from the paper \cite{howard2019searching}}
\label{mv3_rec}
\end{figure}

\textbf{Cosine Learning Rate Decay} As mentioned in text detection, cosine learning rate decay has become the preferred learning rate reduction method. The experiments show that cosine learning rate decay strategy is also effective to enhance the model ability for text recognition.

\textbf{Feature Map Resolution} In order to adapt to multilingual recognition, particularly in Chinese recognition, in PP-OCR the height and width of the CRNN input are set as $32$ and $320$. Then, the strides of the original MobileNetV3 is not appropriate for text recognition. As shown in Figure \ref{mv3_rec}, for the sake of keeping more the horizontal information, we modify the stride of the down sampling feature map except the first one from (2,2) to (2,1). In order to keep more vertical information, we further modify the stride of the second down sampling feature map from (2,1) to (1,1). Thus, the stride of the second down sampling feature map $s_2$ affects the resolution of the whole feature map and the accuracy of the text recognizer dramaticly. In PP-OCR, $s_2$ is set as (1,1) to achieve the better performance empirically.

\textbf{Regularization Parameters} Overfitting is a common term in machine learning. A simple understanding is that the model performs well on the training data, but it performs poorly on the test data. To avoid overfitting, many regular ways have been proposed. Among them, weight\_decay is one of the widely used ways to avoid overfitting. After the final loss function, L2 regularization (L2\_decay) is added to the loss function. With the help of L2 regularization, the weight of the network tend to choose a smaller value, and finally the parameters in the entire network tends to 0, and the generalization performance of the model is improved accordingly. For text recognition, L2\_decay has a great influence on the accuracy.

\textbf{Learning Rate Warm-up} Similar as the text detection, learning rate warm-up is also helping the text recognition. For text recognition, the experiments show that using this strategy is also effective.

\textbf{Light Head} A full connection layer is used to encode the sequence features to the predicted characters in the ordinary. The dimension of the sequence features have an impact on the model size of a text recognizer, especially for Chinese recognition whose characters are more than 6 thousands. Meanwhile, it is not that the higher of the dimension, the stronger of the ability of the sequence features representation. In PP-OCR, the dimension of the sequence features is set to 48 empirically. 

\textbf{Pre-trained Model} If the training data is fewer, fine tune the existing networks, which are trained on a large data set such as ImageNet, to achieve fast convergence and better accuracy. The transfer learning in image classification and object detection show the above strategy is effective. In real scenes, the data used for text recognition is often limited. If the models are trained with tens of millions samples, even if they are synthesized ones, the accuracy can be significantly improved with the above models. We demonstrate the effectiveness of this strategy through experiments. 

\textbf{PACT Quantization} 
We adopt the similar quantization scheme of the direction classification to reduce the model size of a text recognizer except for skipping the LSTM layers. Those layers will not be quantified at present since the complexity of LSTM quantization.

\begin{table*}[h]
\begin{center}
\begin{tabular}{c|c|c|c|c}
\hline
& \multicolumn{3}{c|}{Number of training data} & \makecell[c]{Number of validation data} \\
\cline{2-5}
Task & Total & Real & Synthesis & Real \\
\hline
Text Detection & 97K & 68K & 29K & 500  \\
Direction Classification & 600K & 100K & 500K & 310K  \\
Text Recognition & 17.9M & 1.9M & 16M & 18.7K  \\
\hline
\end{tabular}
\end{center}
\caption{Statistics of dataset for Chinese and English Recognition.}
\label{dataset_ch}
\end{table*}

\begin{table*}[h]
\begin{center}
\begin{tabular}{c|c|c|c|c|c|c|c}
\hline
& & \multicolumn{3}{c|}{Number of training data} & \multicolumn{3}{c}{Number of validation data} \\
\cline{3-8}
Task & \makecell[c]{Character \\ Number} & Total & Real & Synthesis & Total & Real & Synthesis \\
\hline
\makecell[c]{Chinese and English \\ Recognition} & 6622 & 17.9M & 1.9M & 16M & 18.7K & 18.7K & 0 \\
\hline
\makecell[c]{Alphanumeric Symbols \\ Recognition} & 63 & 15M & 0 & 15M & 12K & 12K & 0 \\
\hline
French Recognition & 118 & 1.08M & 0 & 1.08M & 80K & 0 & 80K \\
\hline
Japanese Recognition & 4399 & 0.99M & 0 & 0.99M & 80K & 0 & 80K \\
\hline
Korean Recognition & 3636 & 0.94M & 0 & 0.94M & 80K & 0 & 80K \\
\hline
German Recognition & 131 & 1.96M & 0 & 1.96M & 170K & 0 & 170K \\
\hline
\end{tabular}
\end{center}
\caption{Statistics of dataset for multilingual recognition.}
\label{dataset_mul}
\end{table*}

\section{Experiments}

\subsection{Experimental Setup}

\subsubsection{DataSets}
As shown in Table \ref{dataset_ch}, in order to implement a practical OCR system, we construct a large-scale dataset for Chinese and English recognition as an example.

For text detection, there are 97k training images and 500 validation images. Among the training images, 68K images are real scene images, which come from some public datasets and Baidu image search. The public datasets used include LSVT \cite{sun2019chinese}, RCTW-17 \cite{shi2017icdar2017}, MTWI 2018 \cite{mtwi}, CASIA-10K \cite{he2018multi}, SROIE \cite{huang2019icdar2019}, MLT 2019 \cite{nayef2019icdar2019}, BDI \cite{karatzas2011icdar}, MSRA-TD500 \cite{yao2012detecting} and CCPD 2019 \cite{xu2018towards}. Most the training images from Baidu image search are document text images. The remaining 29K synthetic images mainly focus on the scenarios for long text, multi direction text and table text. All the validation images come from the real scenes.

For direction classification,  there are 600k training images and 310K validation images. Among the training images, 100K images are real scene images, which come from the public datasets (LSVT, RCTW-17, MTWI 2018). They are horizontal text which rectify and crop the ground truth of the images. The remaining 500K synthetic images mainly focus on the reversed text. We use the vertical fonts to synthesize some text images and then rotate them horizontally. All the validation images come from the real scenes.

For text recognition, there are 17.9M training images and 18.7K validation images. Among the training images, 1.9M images are real scene images, which come from some public datasets and Baidu image search. The public datasets used include LSVT, RCTW-17, MTWI 2018 and CCPD 2019. The remaining 16M synthetic images mainly focus on the scenarios for different backgrounds, translation, rotation, perspective transformation, line disturb, noise, vertical text and so on. The corpus of synthetic images come from the real scene images. All the validation images also come from the real scenes.

In order to conduct ablation experiments quickly and choose the appropriate strategies, we select 4k images from the real scene training images for text detection, and 300k ones from the real scene training images for text recognition.

In addition, we collected 300 images for different real application scenarios to evaluate the overall OCR system, including contract samples, license plates, nameplates, train tickets, test sheets, forms, certificates, street view images, business cards, digital meter, etc. Figure \ref{ex_SRT} and Figure \ref{ex_doc} show some images of the test set.

Furthermore, to verify the proposed PP-OCR for other languages, we also collect some corpus for alphanumeric symbols recognition, French recognition, Korean recognition, Japanese recognition and German recognition. Then synthesize the text line images for text recognition. Some images for alphanumeric symbols recognition come from the public datasets, ST \cite{gupta2016synthetic} and SRN \cite{yu2020towards}. Table \ref{dataset_mul} shows the statistics. Since MLT 2019 for text detection includes multilingual images, the text detector for Chinese and English recognition also can support multi language text detection. Due to the limited data, we haven't found the proper data to train the direction classifier for multilingual.

The data synthesis tool used in text detection and text recognition is modified from text\_render \cite{textrender}.

\subsubsection{Implementation Details}

\begin{table*}[h]
\begin{center}
\begin{tabular}{c|c|c|c|c|c|c|c|c}
\hline
\makecell[c]{inner\_channel \\ of the head} & \makecell[c]{Remove \\ SE} & \makecell[c]{Cosine \\ Learning \\ Rate Decay} & \makecell[c]{Learning \\ Rate \\ Warm-up} & Precision & Recall & HMean & \makecell[c]{Model \\ Size (M)} & \makecell[c]{Inference Time \\ (CPU, ms)} \\
\hline
256 &  &  &  & 0.6821 & 0.5560 & 0.6127 & 7 & 406  \\
96 &  &  &  & 0.6677 & 0.5524 & 0.6046 & 4.1 & 213  \\
96 & $\surd$ &  &  & 0.6952 & 0.5413 & 0.6087 & 2.6 & 173  \\
96 & $\surd$ & $\surd$ &  & 0.7034 & 0.5404 & 0.6112 & 2.6 & 173  \\
96 & $\surd$ & $\surd$ & $\surd$ & 0.7349 & 0.5420 & 0.6239 & 2.6 & 173  \\
\hline
\end{tabular}
\end{center}
\caption{Ablation study of inner\_channel of the head, SE, cosine learning rate decay, learning rate warm-up for text detection.}
\label{det_tricks}
\end{table*}

\begin{table}[h]
\begin{center}
\begin{tabular}{c|c|c|c|c}
\hline
& & & \multicolumn{2}{c}{Number of Epochs} \\
\cline{4-5}
Task & \makecell[c]{Initial \\ Learning \\ Rate} & \makecell[c]{Batch \\ Size} & \makecell[c]{Ablation \\ Data}  & \makecell[c]{Total \\ Data} \\
\hline
\makecell[c]{Text \\ Detection} & 0.001 & 16 & 400 & 60  \\
\hline
\makecell[c]{Direction \\ Classification} & 0.001 & 512 & 100 & 100  \\
\hline
\makecell[c]{Text \\ Recognition} & 0.001 & 1024 & 500 & 100  \\
\hline
\end{tabular}
\end{center}
\caption{Implementation details of the model training.}
\label{imp_train}
\end{table}

\begin{table}[h]
\begin{center}
\begin{tabular}{c|c|c|c}
\hline
Backbone & HMean & \makecell[c]{Model \\ Size (M)} & \makecell[c]{Inference Time \\ (CPU, ms)} \\
\hline
\makecell[c]{MobileNetV3\_ \\ large\_x1} & 0.6463 & 16 & 447  \\
\hline
\makecell[c]{MobileNetV3\_ \\ large\_x0.5} & 0.6127 & 7 & 406  \\
\hline
\makecell[c]{MobileNetV3\_ \\ large\_x0.35} & 0.5935 & 5.4 & 367  \\
\hline
\makecell[c]{MobileNetV3\_ \\ small\_x1} & 0.5919 & 7.5 & 380  \\
\hline
\end{tabular}
\end{center}
\caption{Compare the performance of the different backbones for text detection.}
\label{det_bockbone}
\end{table}

\begin{table}[h]
\begin{center}
\begin{tabular}{c|c|c|c}
\hline
\makecell[c]{FPGM \\ Pruner} & HMean & \makecell[c]{Model \\ Size (M)} & \makecell[c]{Inference Time \\ (SD 855, ms)} \\
\hline
 & 0.6239 & 2.6 & 164  \\
$\surd$ & 0.6169 & 1.4 & 133  \\
\hline
\end{tabular}
\end{center}
\caption{Ablation study of FPGM pruner for text detection.}
\label{det_pruner}
\end{table}

\begin{table}[h]
\begin{center}
\begin{tabular}{c|c|c|c}
\hline
Backbone & Accuracy & \makecell[c]{Model \\ Size (M)} & \makecell[c]{Inference \\ Time \\ (CPU, ms)} \\
\hline
\makecell[c]{MobileNetV3\_ \\ small\_x0.5} & 0.9494 & 1.34 & 3.22  \\
\hline
\makecell[c]{MobileNetV3\_ \\ small\_x0.35} & 0.9403 & 0.85 & 3.21  \\
\hline
\makecell[c]{ShuffleNetV2\_ \\ x0.5} & 0.9017 & 1.72 & 3.41  \\
\hline
\end{tabular}
\end{center}
\caption{Compares the performance of the different backbones for direction classification.}
\label{dir_backbone}
\end{table}

We use Adam optimizer to train all the models and adopt cosine learning rate decay as the learning rate schedule. The initial learning rate, batch size and the number of epochs for different tasks can be found in Table \ref{imp_train}. When we obtain the trained models, FPGM pruner and PACT quantization can be used to reduce the model size further with the above models as the pre-trained 
ones. The training processes of FPGM pruner and PACT quantization are similar as previous.

In the inference period, HMean is used to evaluate the performance of a text detector. Accuracy is used to evaluate the performance of a direction classifier or a text recognizer. F-score is used to evaluate the performance of an OCR system. In order to calculate F-score, a correct text recognition result should be the accurate location and the same text. GPU inference time is tested on a single T4 GPU. CPU inference time is tested on a Intel(R) Xeon(R) Gold 6148. We use the Snapdragon 855 (SD 855) to evaluate the inference time of the quantification models.

\subsection{Text Detection}

Table \ref{det_bockbone} compares the performance of the different backbones for text detection. HMean, the model size and the inference time of the different scales of MobileNetV3 change greatly. In PP-OCR, we choose MobileNetV3\_large\_x0.5 to balance accuracy and efficiency. 

Tabel \ref{det_tricks} shows the ablation study of inner\_channel of the head, SE, cosine learning rate decay, learning rate warm-up for text detection. Firstly, by reducing the internal channels of the detector head from 256 to 96, the model size was reduced by 41\%, and the inference time was accelerated by nearly 50\% with HMean only dropped slightly. Therefore, reducing the inner channel is an effective way to lighten the detector. Then, when remove the SE block of the detector backbone, the model size is reduced 36.6\% and the inference time has accelerated 18.8\% further. Meanwhile, HMean will not be affected. Therefore, for text detection, the accuracy improvement of SE blocks is limited, but the time cost is very high. Finally, using both cosine learning rate decay instead of the fix learning rate and learning rate warm-up, HMean will be improved obviously. At the same time, the model size and the inference time will not be affected. Cosine learning rate decay and learning rate warm-up are effective strategies for text detection.

Table \ref{det_pruner} shows the ablation study of FPGM pruner for text detection. Using FPGM pruner, the model size is reduced 46.2\% and the inference time has accelerated 18.9\% on SD 855 device with HMean slightly dropped. Therefore, FPGM pruner can prune the text detection model effectively.

\begin{table*}[h]
\begin{center}
\begin{tabular}{c|c|c|c|c}
\hline
\makecell[c]{Input Resolution} & \makecell[c]{PACT Quantization} & Accuracy & \makecell[c]{Model Size (M)} & \makecell[c]{Inference Time (SD 855, ms)} \\
\hline
$3\times32\times100$ &  & 0.9212 & 0.85 & 3.19  \\
$3\times48\times192$ &  & 0.9403 & 0.85 & 3.21  \\
$3\times48\times192$ & $\surd$ & 0.9456 & 0.46 & 2.38  \\
\hline
\end{tabular}
\end{center}
\caption{Ablation study of input resolution and PACT quantization for direction classification.}
\label{dir_qu}
\end{table*}

\begin{table}[h]
\begin{center}
\begin{tabular}{c|c}
\hline
Data Augmentation & Accuracy \\
\hline
NO & 0.8879  \\
BDA & 0.9134  \\
BDA+CutMix & 0.9083  \\
BDA+Mixup & 0.9104  \\
BDA+Cutout & 0.9081  \\
BDA+HideAndSeek & 0.8598  \\
BDA+GridMask & 0.9140  \\
BDA+RandomErasing & 0.9193  \\
BDA+AutoAugment & 0.9133  \\
BDA+RandAugment & 0.9212  \\
\hline
\end{tabular}
\end{center}
\caption{Ablation study of data augmentation for direction classification.}
\label{dir_aug}
\end{table}

\begin{table}[h]
\begin{center}
\begin{tabular}{c|c|c|c}
\hline
Backbone & Accuracy & \makecell[c]{Model \\ Size (M)} & \makecell[c]{Inference \\ Time \\ (CPU, ms)} \\
\hline
\makecell[c]{MobileNetV3\_ \\ small\_x0.35} & 0.6288 & 22 & 17  \\
\hline
\makecell[c]{MobileNetV3\_ \\ small\_x0.5} & 0.6556 & 23 & 17.27  \\
\hline
\makecell[c]{MobileNetV3\_ \\ small\_x1} & 0.6933 & 28 & 19.15  \\
\hline
\end{tabular}
\end{center}
\caption{Compares the performance of the different backbones for text recognition. The number of channel in the head is 256.}
\label{rec_bockbone}
\end{table}

\begin{table}[h]
\begin{center}
\begin{tabular}{c|c|c|c}
\hline
\makecell[c]{the number \\ of channel} & Accuracy & \makecell[c]{Model  \\ Size (M)} & \makecell[c]{Inference \\ Time \\ (CPU, ms)} \\
\hline
256 & 0.6556 & 23 & 17.27  \\
\hline
96 & 0.6673 & 8 & 13.36  \\
\hline
64 & 0.6642 & 5.6 & 12.64  \\
\hline
48 & 0.6581 & 4.6 & 12.26  \\
\hline
\end{tabular}
\end{center}
\caption{Ablation study of the number of channel in the head for text recognition. The data augmentation is only used BDA.}
\label{rec_head}
\end{table}

\subsection{Direction Classification}

Table \ref{dir_backbone} compares the performance of different backbones for direction classification. The accuracy of MobileNetV3 with difference scales $(0.35, 0.5)$ are close. The model size and the inference time of MobileNetV3\_small\_x0.35 are much better. Besides, ShuffleNetV2 is used to train a direction classifier in some previous work. From the table, whether it's accuracy or the model size or the inference time, ShuffleNetV2 is not a good choice.

Tabel \ref{dir_aug} shows the ablation study of data augmentation for direction classification. The baseline accuracy of text director classify without data augmentation is only 88.79\%. When we adopt BDA (base data augmentation), the accuracy can boost 2.55\%. We also verified that RandomErasing and RandAugment are useful for text direction classification. Therefore, in PP-OCR, we use BDA (base data augmentation) and RandAugment to train a direction classifier.

Table \ref{dir_qu} shows the ablation study of input resolution and PACT quantization for direction classification. When the input resolution is adjusted from $3\times32\times100$ to $3\times48\times192$, The classification accuracy has improved but the prediction speed is basically unchanged. Furthermore, we also verified quantization strategy is effective in accelerating the prediction speed of the text direction classifier. The model size is reduced 45.9\% and the inference time has accelerated 25.86\%. Accuracy is slight promotion. 

\subsection{Text Recognition}

Table \ref{rec_bockbone} compares the performance of the different backbones for text recognition. The accuracy, the model size and the inference time of the different scales of MobileNetV3 change greatly. In PP-OCR, we choose MobileNetV3\_small\_x0.5 to balance accuracy and efficiency.

\begin{table*}[h]
\begin{center}
\begin{tabular}{c|c|c|c|c|c|c|c}
\hline
Strategy & \makecell[c]{Data \\ Augmentation} & \makecell[c]{Cosine Learning \\ Rate Decay} & Stride & L2\_decay & \makecell[c]{Learning Rate \\ Warm-up} & Accuracy & \makecell[c]{Inference Time \\ (CPU, ms)} \\
\hline
S1 & NO & & (2,1) & 0 &  & 0.5193 & 11.84  \\
S2 & BDA &  & (2,1) & 0 &  & 0.5505 & 11.84  \\
S3 & BDA & $\surd$ & (2,1) & 0 &  & 0.5652 & 11.84  \\
S4 & BDA & $\surd$ & (1,1) & 0 &  & 0.6179 & 12.96  \\
S5 & BDA & $\surd$ & (1,1) & $1e-5$ &  & 0.6519 & 12.96  \\
S6 & BDA & $\surd$ & (1,1) & $1e-5$ & $\surd$ & 0.6581 & 12.96  \\
S7 & BDA+TIA & $\surd$ & (1,1) & $1e-5$ & $\surd$ & 0.6670 & 12.96  \\
\hline
\end{tabular}
\end{center}
\caption{Ablation study of data augmentation, cosine learning rate decay, the stride of the second down sampling feature map, regularization parameters L2\_decay and learning rate warm-up for text recognition. Backbone is MobileNetV3\_small\_x0.5. The number of channel in the head is 48.}
\label{rec_tricks}
\end{table*}

\begin{table}[h]
\begin{center}
\begin{tabular}{c|c|c|c}
\hline
\makecell[c]{PACT \\ Quantization} & Accuracy & \makecell[c]{Model \\ Size (M)} & \makecell[c]{Inference \\ Time \\ (SD 855, ms)} \\
\hline
 & 0.6581 & 4.6 & 12  \\
\hline
 $\surd$ & 0.674 & 1.5 & 11  \\
\hline
\end{tabular}
\end{center}
\caption{Ablation study of PACT quantization for text recognition.}
\label{rec_quan}
\end{table}

\begin{table}[h]
\begin{center}
\begin{tabular}{c|c|c|c}
\hline
\makecell[c]{Slim} & \makecell[c]{F-score} & \makecell[c]{Model \\ Size (M)} & \makecell[c]{Inference Time \\ (SD 855, ms)} \\
\hline
 & 0.5193 & 8.1 & 306 \\ 
\hline
 $\surd$ & 0.5210 & 3.5 & 268 \\
\hline
\end{tabular}
\end{center}
\caption{Ablation study of the prunner or quantization for the OCR system.}
\label{sys_quan}
\end{table}

\begin{table}[h]
\begin{center}
\begin{tabular}{c|c|c|c|c}
\hline
 & & & \multicolumn{2}{c}{Inference Time (ms)} \\
\cline{4-5}
\makecell[c]{Model \\ Type} & \makecell[c]{F-\\score} & \makecell[c]{Model \\ Size (M)} & \makecell[c]{CPU} & \makecell[c]{T4 GPU}\\
\hline
\makecell[c]{Ultra \\ lightweight} & 0.5193 & 8.1 & 421 & 137 \\ 
\hline
\makecell[c]{Large \\ scale} & 0.5414 & 155.1 & 1199 & 204 \\
\hline
\end{tabular}
\end{center}
\caption{Compare between the ultra lightweight OCR system and the large scale one.}
\label{sys_scale}
\end{table}

\begin{figure*}[t]
\centering
\subfigure{
\centering
\includegraphics[width=13cm]{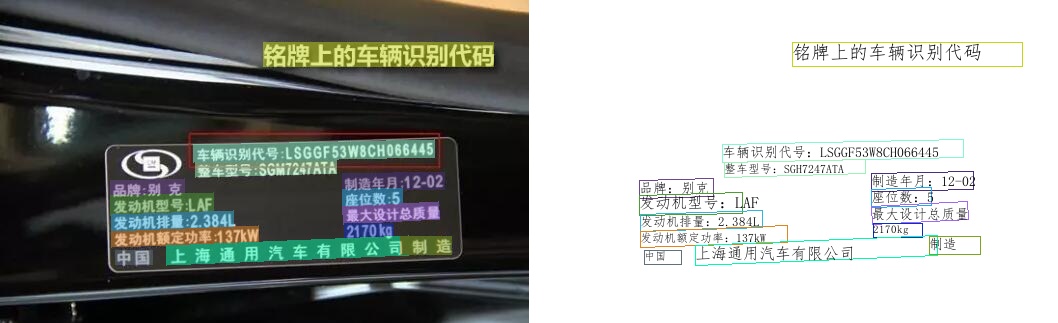}
}

\subfigure{
\centering
\includegraphics[width=13cm]{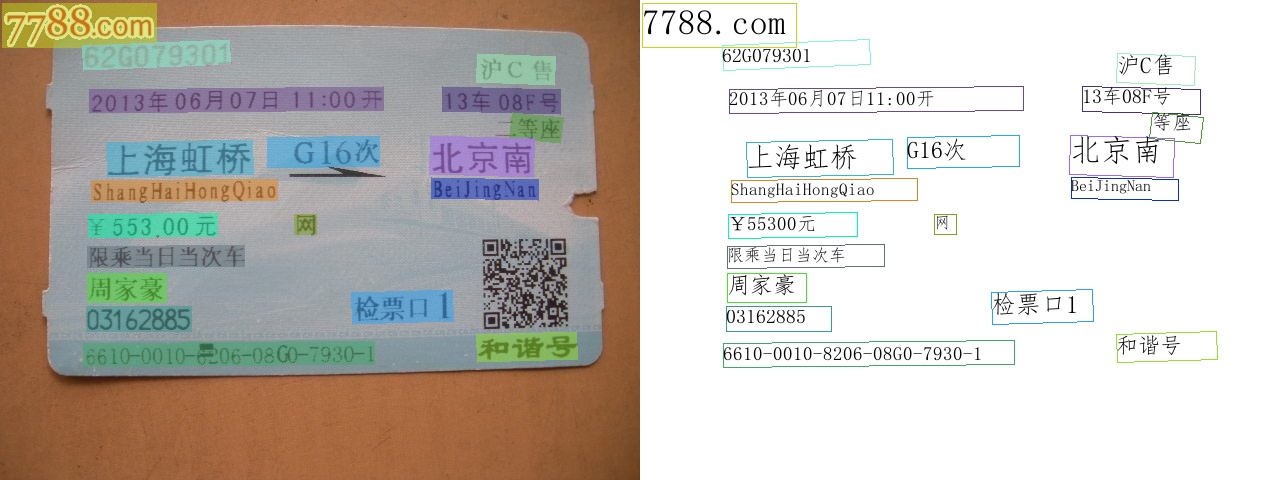}
}

\subfigure{
\centering
\includegraphics[width=13cm]{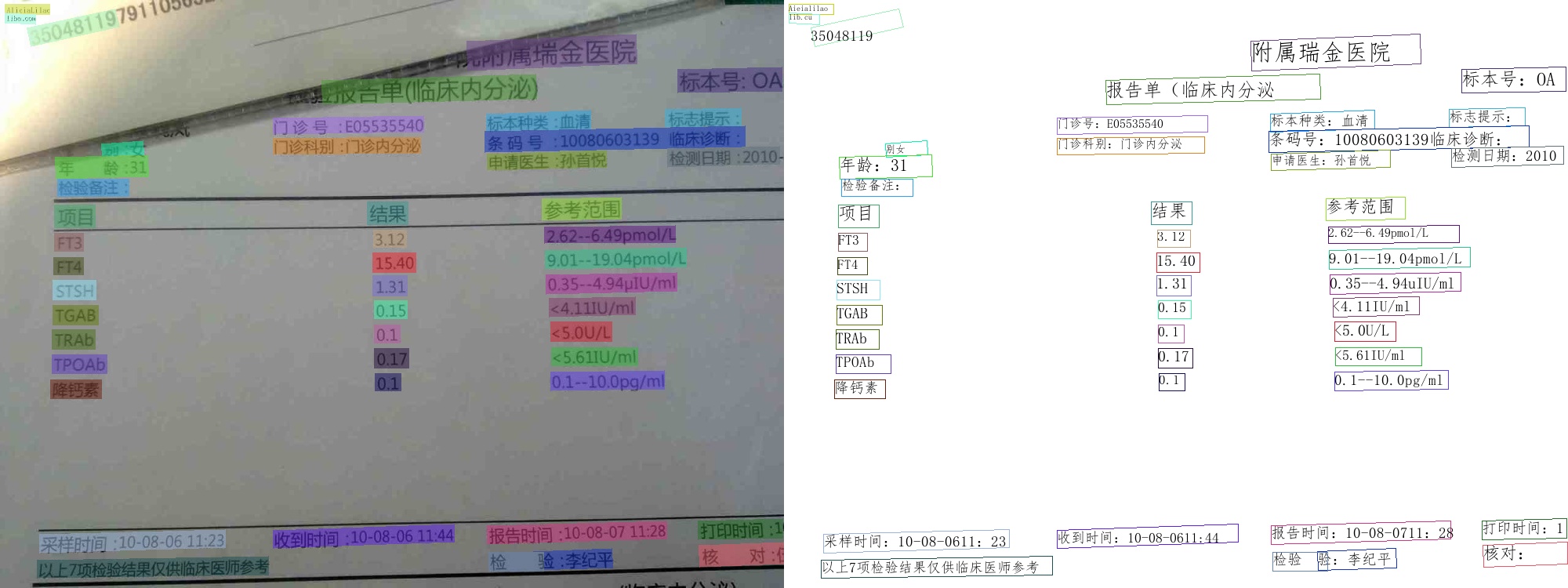}
}

\caption{Some image results of the proposed PP-OCR system for Chinese and English recognition.}
\label{res_p2}
\end{figure*}

\begin{figure*}[t]
\centering

\subfigure{
\centering
\includegraphics[width=12cm]{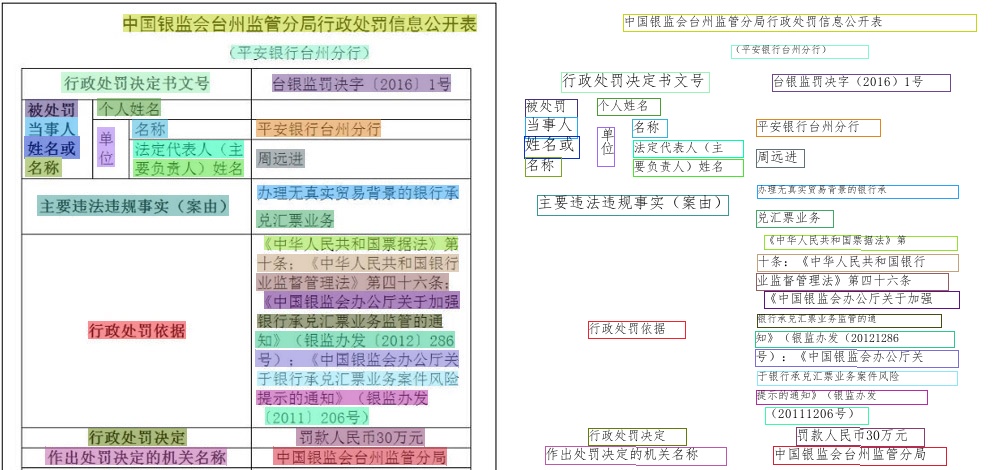}
}

\subfigure{
\centering
\includegraphics[width=11cm]{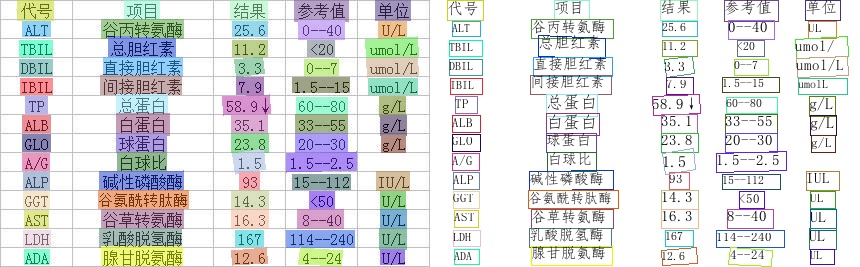}
}

\subfigure{
\centering
\includegraphics[width=10cm]{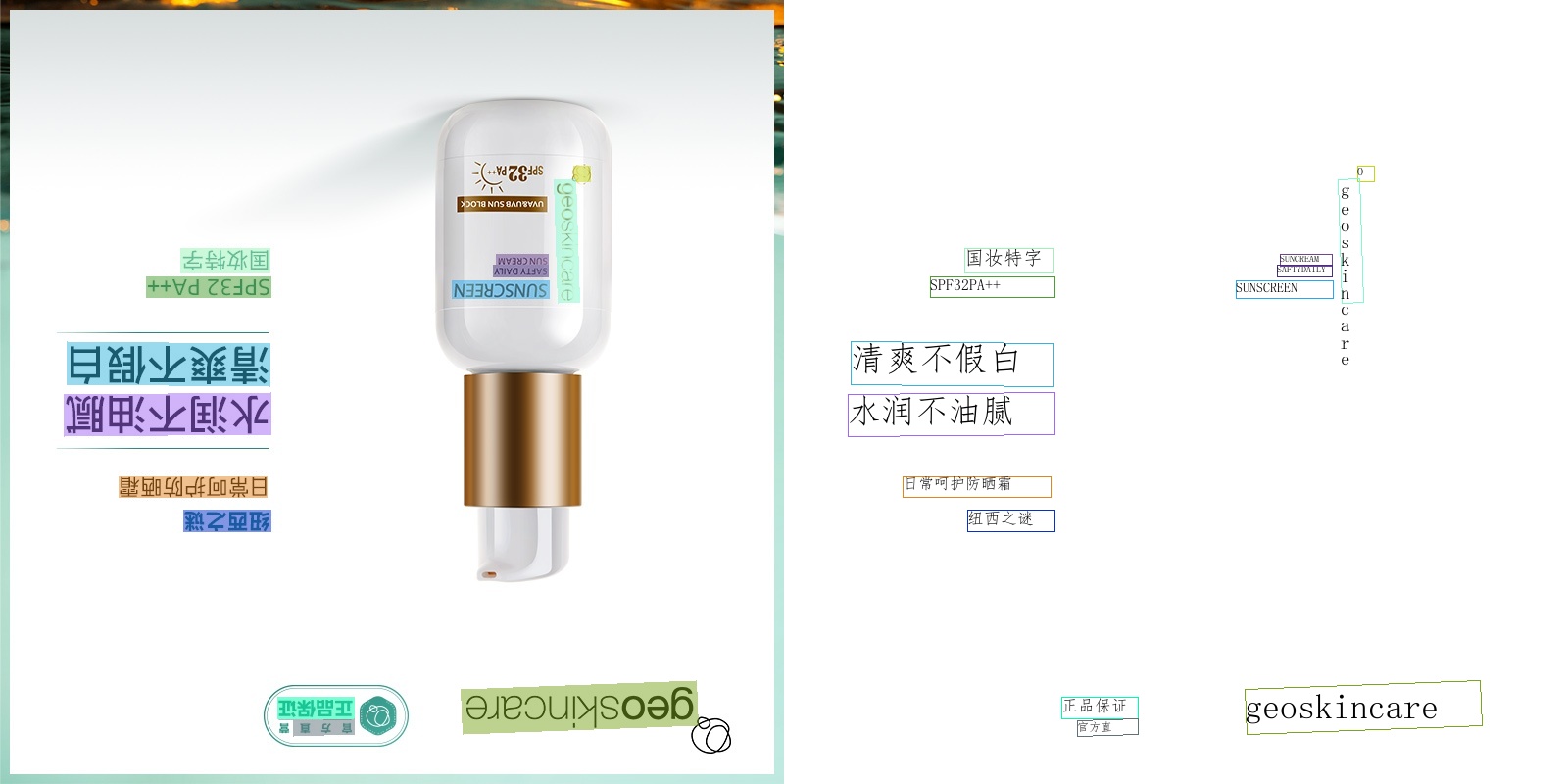}
}

\caption{Some image results of the proposed PP-OCR system for Chinese and English recognition.}
\label{res_p4}
\end{figure*}


\begin{figure*}[t]
\centering
\subfigure{
\centering
\includegraphics[width=11cm]{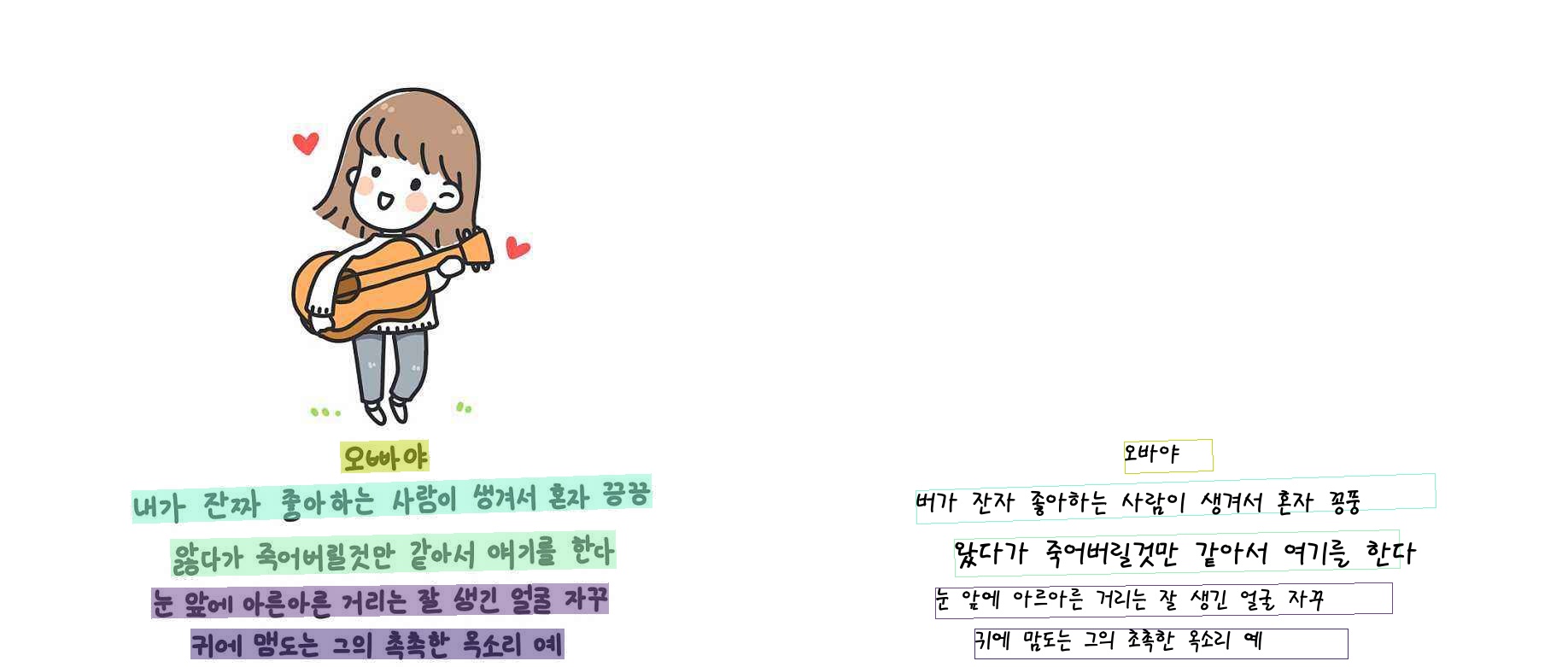}
}

\subfigure{
\centering
\includegraphics[width=10cm]{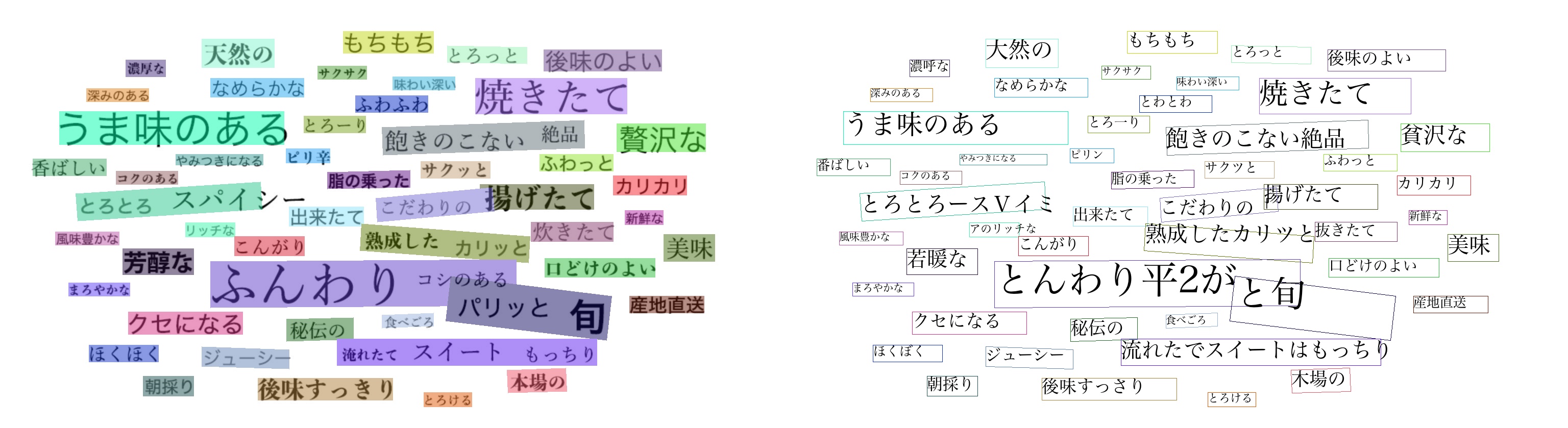}
}

\subfigure{
\centering
\includegraphics[width=8cm]{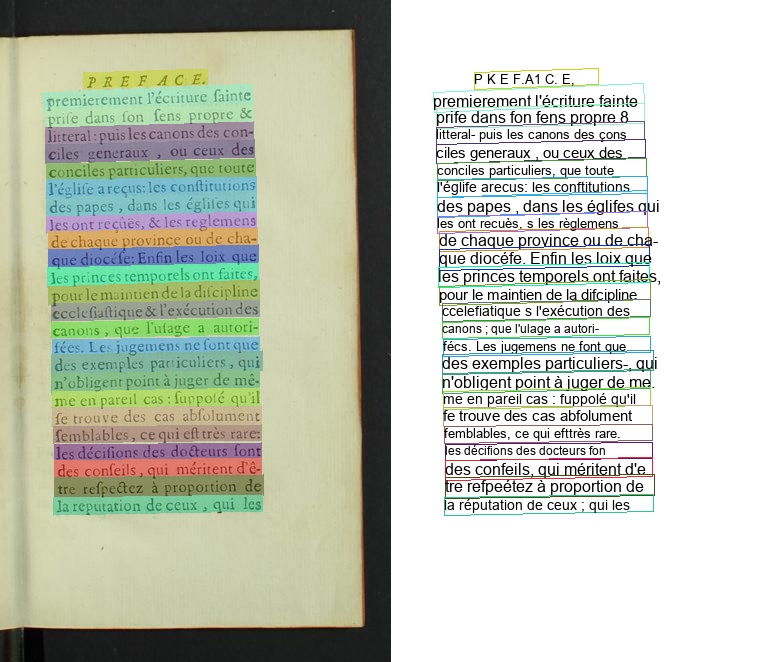}
}

\caption{Some image results of the proposed PP-OCR system for multilingual recognition.}
\label{res_p3}
\end{figure*}

Table \ref{rec_head} compares the number of channel in the CRNN head for text recognition. Reduce the number of channel from 256 to 48, the model size is reduced from 23M to 4.6M and the inference time has accelerated nearly 30\%. However, the accuracy will not be affected. We can see the number of channel in the head has a great influence on the model size of a lightweight text recognizer.

Tabel \ref{rec_tricks} shows the ablation study of data augmentation, cosine learning rate decay, the stride of the second down sampling feature map, regularization parameters L2\_decay and learning rate warm-up for text recognition.

To verify the advantages of each strategy, the setting of the basic experimental is the strategy S1. When using BDA, the accuracy will be improved 3.12\%. Data augmentation is very necessary for text recognition. When we adopt the cosine learning rate decay further, the accuracy will be improved 1.47\%. The cosine learning rate is an effective strategy for text recognition. Next, when we increase the feature map resolution and reduce the stride of the second down sampling feature map from (2,1) to (1,1), the accuracy will be improved 5.27\%. Then, when we adjust the regularization parameters L2\_decay from 0 to $1e-5$ further, the accuracy will be improved 3.4\%. The feature map resolution and L2\_decay have a great influence on the performance. Final, using learning rate warm-up, the accuracy will be improved 0.62\%. Using TIA data augmentation, the accuracy will be improved 0.91\%. Learning rate warm-up and TIA also are effective strategies for text recognition.

Tabel \ref{rec_quan} shows the ablation study of PACT quantization for text recognition. When we use PACT quantization, the model size is reduced 67.39\% and the inference time has accelerated 8.3\%. Since there was no quantification on LSTM, The acceleration is not obvious. However, accuracy achieves a significant improvement. Therefore, PACT quantization also is an effective strategy for reducing the model size of a text recognizer.

In the end, we will illustrate the effect of pre-trained model. We utilize 17.9M training images to learn a text recognizer. Then, use this model as the pre-trained model to fine-tuning the samples for the ablation experiments. When using above pre-trained model, the accuracy will go from 65.81\% to 69\% and the effect is very obvious.

\subsection{System Performance}

Table \ref{sys_quan} shows the ablation study of the prunner or quantization for the OCR system.  When we use the slim approaches, the model size is reduced 55.7\% and the inference time has accelerated 12.42\%. F-score has no impact. The inference time includes pre-process and post-process of each parts of the system. Therefore, FPGM pruner and PACT quantization also are effective strategies for reducing the model size.

To compare the gap between the proposed ultra lightweight OCR system and large-scale OCR system, we also train a large-scale OCR system and use Res18\_vd as the text detector backbone and Res34\_vd as the text recognizer backbone. Table \ref{sys_scale} shows the comparison. F-score of the large-scale OCR system is higher than the ultra lightweight OCR system, but the model size and the inference time of the ultra lightweight system are better obviously.   

Figure \ref{res_p2} and Figure \ref{res_p4} show some image results of the proposed PP-OCR system for Chinese and English recognition. Figure \ref{res_p3} show some image results of the proposed PP-OCR system for multilingual recognition.

\section{Conclusions}
In this paper, we propose a practical ultra lightweight OCR system, PP-OCR, which the overall model size is only 3.5M for recognizing 6622 Chinese characters and 2.8M for recognizing 63 alphanumeric symbols. We introduce a bag of strategies to either enhance the model ability or light the model. The corresponding ablation experiments are also provided. Meanwhile, some practical ultra lightweight OCR models are released with a large-scale dataset.

\bibstyle{aaai21}
\bibliography{eg}

\begin{thebibliography}{30}
\providecommand{\natexlab}[1]{#1}
\providecommand{\url}[1]{\texttt{#1}}
\providecommand{\urlprefix}{URL }
\expandafter\ifx\csname urlstyle\endcsname\relax
  \providecommand{\doi}[1]{doi:\discretionary{}{}{}#1}\else
  \providecommand{\doi}{doi:\discretionary{}{}{}\begingroup
  \urlstyle{rm}\Url}\fi

\bibitem[{Chen(2020)}]{chen2020gridmask}
Chen, P. 2020.
\newblock GridMask data augmentation.
\newblock \emph{arXiv preprint arXiv:2001.04086} .

\bibitem[{Choi et~al.(2018)Choi, Wang, Venkataramani, Chuang, Srinivasan, and
  Gopalakrishnan}]{choi2018pact}
Choi, J.; Wang, Z.; Venkataramani, S.; Chuang, P. I.-J.; Srinivasan, V.; and
  Gopalakrishnan, K. 2018.
\newblock Pact: Parameterized clipping activation for quantized neural
  networks.
\newblock \emph{arXiv preprint arXiv:1805.06085} .

\bibitem[{Cubuk et~al.(2019)Cubuk, Zoph, Mane, Vasudevan, and
  Le}]{cubuk2019autoaugment}
Cubuk, E.~D.; Zoph, B.; Mane, D.; Vasudevan, V.; and Le, Q.~V. 2019.
\newblock Autoaugment: Learning augmentation strategies from data.
\newblock In \emph{Proceedings of the IEEE conference on computer vision and
  pattern recognition}, 113--123.

\bibitem[{Cubuk et~al.(2020)Cubuk, Zoph, Shlens, and Le}]{cubuk2020randaugment}
Cubuk, E.~D.; Zoph, B.; Shlens, J.; and Le, Q.~V. 2020.
\newblock Randaugment: Practical automated data augmentation with a reduced
  search space.
\newblock In \emph{Proceedings of the IEEE/CVF Conference on Computer Vision
  and Pattern Recognition Workshops}, 702--703.

\bibitem[{DeVries and Taylor(2017)}]{devries2017improved}
DeVries, T.; and Taylor, G.~W. 2017.
\newblock Improved regularization of convolutional neural networks with cutout.
\newblock \emph{arXiv preprint arXiv:1708.04552} .

\bibitem[{Gupta, Vedaldi, and Zisserman(2016)}]{gupta2016synthetic}
Gupta, A.; Vedaldi, A.; and Zisserman, A. 2016.
\newblock Synthetic data for text localisation in natural images.
\newblock In \emph{Proceedings of the IEEE conference on computer vision and
  pattern recognition}, 2315--2324.

\bibitem[{He and Yang(2018)}]{mtwi}
He, M.; and Yang, Z. 2018.
\newblock ICPR 2018 contest on robust reading for multi-type web images (MTWI).
\newblock
  \url{https://tianchi.aliyun.com/competition/entrance/231651/information}.

\bibitem[{He et~al.(2019{\natexlab{a}})He, Zhang, Zhang, Zhang, Xie, and
  Li}]{he2019bag}
He, T.; Zhang, Z.; Zhang, H.; Zhang, Z.; Xie, J.; and Li, M.
  2019{\natexlab{a}}.
\newblock Bag of tricks for image classification with convolutional neural
  networks.
\newblock In \emph{Proceedings of the IEEE Conference on Computer Vision and
  Pattern Recognition}, 558--567.

\bibitem[{He et~al.(2018)He, Zhang, Yin, and Liu}]{he2018multi}
He, W.; Zhang, X.-Y.; Yin, F.; and Liu, C.-L. 2018.
\newblock Multi-oriented and multi-lingual scene text detection with direct
  regression.
\newblock \emph{IEEE Transactions on Image Processing} 27(11): 5406--5419.

\bibitem[{He et~al.(2019{\natexlab{b}})He, Liu, Wang, Hu, and
  Yang}]{he2019filter}
He, Y.; Liu, P.; Wang, Z.; Hu, Z.; and Yang, Y. 2019{\natexlab{b}}.
\newblock Filter pruning via geometric median for deep convolutional neural
  networks acceleration.
\newblock In \emph{Proceedings of the IEEE Conference on Computer Vision and
  Pattern Recognition}, 4340--4349.

\bibitem[{Howard et~al.(2019)Howard, Sandler, Chu, Chen, Chen, Tan, Wang, Zhu,
  Pang, Vasudevan et~al.}]{howard2019searching}
Howard, A.; Sandler, M.; Chu, G.; Chen, L.-C.; Chen, B.; Tan, M.; Wang, W.;
  Zhu, Y.; Pang, R.; Vasudevan, V.; et~al. 2019.
\newblock Searching for mobilenetv3.
\newblock In \emph{Proceedings of the IEEE International Conference on Computer
  Vision}, 1314--1324.

\bibitem[{Hu, Shen, and Sun(2018)}]{hu2018squeeze}
Hu, J.; Shen, L.; and Sun, G. 2018.
\newblock Squeeze-and-excitation networks.
\newblock In \emph{Proceedings of the IEEE conference on computer vision and
  pattern recognition}, 7132--7141.

\bibitem[{Huang et~al.(2019)Huang, Chen, He, Bai, Karatzas, Lu, and
  Jawahar}]{huang2019icdar2019}
Huang, Z.; Chen, K.; He, J.; Bai, X.; Karatzas, D.; Lu, S.; and Jawahar, C.
  2019.
\newblock Icdar2019 competition on scanned receipt ocr and information
  extraction.
\newblock In \emph{2019 International Conference on Document Analysis and
  Recognition (ICDAR)}, 1516--1520. IEEE.

\bibitem[{Karatzas et~al.(2011)Karatzas, Mestre, Mas, Nourbakhsh, and
  Roy}]{karatzas2011icdar}
Karatzas, D.; Mestre, S.~R.; Mas, J.; Nourbakhsh, F.; and Roy, P.~P. 2011.
\newblock ICDAR 2011 robust reading competition-challenge 1: reading text in
  born-digital images (web and email).
\newblock In \emph{2011 International Conference on Document Analysis and
  Recognition}, 1485--1490. IEEE.

\bibitem[{Li et~al.(2016)Li, Kadav, Durdanovic, Samet, and
  Graf}]{li2016pruning}
Li, H.; Kadav, A.; Durdanovic, I.; Samet, H.; and Graf, H.~P. 2016.
\newblock Pruning filters for efficient convnets.
\newblock \emph{arXiv preprint arXiv:1608.08710} .

\bibitem[{Liao et~al.(2020)Liao, Wan, Yao, Chen, and Bai}]{liao2020real}
Liao, M.; Wan, Z.; Yao, C.; Chen, K.; and Bai, X. 2020.
\newblock Real-Time Scene Text Detection with Differentiable Binarization.
\newblock In \emph{AAAI}, 11474--11481.

\bibitem[{Lin et~al.(2017)Lin, Doll{\'a}r, Girshick, He, Hariharan, and
  Belongie}]{lin2017feature}
Lin, T.-Y.; Doll{\'a}r, P.; Girshick, R.; He, K.; Hariharan, B.; and Belongie,
  S. 2017.
\newblock Feature pyramid networks for object detection.
\newblock In \emph{Proceedings of the IEEE conference on computer vision and
  pattern recognition}, 2117--2125.

\bibitem[{Luo et~al.(2020)Luo, Zhu, Jin, and Wang}]{luo2020learn}
Luo, C.; Zhu, Y.; Jin, L.; and Wang, Y. 2020.
\newblock Learn to Augment: Joint Data Augmentation and Network Optimization
  for Text Recognition.
\newblock In \emph{Proceedings of the IEEE/CVF Conference on Computer Vision
  and Pattern Recognition}, 13746--13755.

\bibitem[{Nayef et~al.(2019)Nayef, Patel, Busta, Chowdhury, Karatzas, Khlif,
  Matas, Pal, Burie, Liu et~al.}]{nayef2019icdar2019}
Nayef, N.; Patel, Y.; Busta, M.; Chowdhury, P.~N.; Karatzas, D.; Khlif, W.;
  Matas, J.; Pal, U.; Burie, J.-C.; Liu, C.-l.; et~al. 2019.
\newblock ICDAR2019 robust reading challenge on multi-lingual scene text
  detection and recognition—RRC-MLT-2019.
\newblock In \emph{2019 International Conference on Document Analysis and
  Recognition (ICDAR)}, 1582--1587. IEEE.

\bibitem[{Sanster(2018)}]{textrender}
Sanster. 2018.
\newblock Generate text images for training deep learning ocr model.
\newblock \url{https://github.com/Sanster/text_renderer}.

\bibitem[{Shi, Bai, and Yao(2016)}]{shi2016end}
Shi, B.; Bai, X.; and Yao, C. 2016.
\newblock An end-to-end trainable neural network for image-based sequence
  recognition and its application to scene text recognition.
\newblock \emph{IEEE transactions on pattern analysis and machine intelligence}
  39(11): 2298--2304.

\bibitem[{Shi et~al.(2017)Shi, Yao, Liao, Yang, Xu, Cui, Belongie, Lu, and
  Bai}]{shi2017icdar2017}
Shi, B.; Yao, C.; Liao, M.; Yang, M.; Xu, P.; Cui, L.; Belongie, S.; Lu, S.;
  and Bai, X. 2017.
\newblock ICDAR2017 competition on reading chinese text in the wild (RCTW-17).
\newblock In \emph{2017 14th IAPR International Conference on Document Analysis
  and Recognition (ICDAR)}, volume~1, 1429--1434. IEEE.

\bibitem[{Singh and Lee(2017)}]{singh2017hide}
Singh, K.~K.; and Lee, Y.~J. 2017.
\newblock Hide-and-seek: Forcing a network to be meticulous for
  weakly-supervised object and action localization.
\newblock In \emph{2017 IEEE international conference on computer vision
  (ICCV)}, 3544--3553. IEEE.

\bibitem[{Sun et~al.(2019)Sun, Liu, Liu, Han, Ding, and Liu}]{sun2019chinese}
Sun, Y.; Liu, J.; Liu, W.; Han, J.; Ding, E.; and Liu, J. 2019.
\newblock Chinese street view text: Large-scale chinese text reading with
  partially supervised learning.
\newblock In \emph{Proceedings of the IEEE International Conference on Computer
  Vision}, 9086--9095.

\bibitem[{Xu et~al.(2018)Xu, Yang, Meng, Lu, Huang, Ying, and
  Huang}]{xu2018towards}
Xu, Z.; Yang, W.; Meng, A.; Lu, N.; Huang, H.; Ying, C.; and Huang, L. 2018.
\newblock Towards end-to-end license plate detection and recognition: A large
  dataset and baseline.
\newblock In \emph{Proceedings of the European conference on computer vision
  (ECCV)}, 255--271.

\bibitem[{Yao et~al.(2012)Yao, Bai, Liu, Ma, and Tu}]{yao2012detecting}
Yao, C.; Bai, X.; Liu, W.; Ma, Y.; and Tu, Z. 2012.
\newblock Detecting texts of arbitrary orientations in natural images.
\newblock In \emph{2012 IEEE conference on computer vision and pattern
  recognition}, 1083--1090. IEEE.

\bibitem[{Yu et~al.(2020)Yu, Li, Zhang, Liu, Han, Liu, and
  Ding}]{yu2020towards}
Yu, D.; Li, X.; Zhang, C.; Liu, T.; Han, J.; Liu, J.; and Ding, E. 2020.
\newblock Towards accurate scene text recognition with semantic reasoning
  networks.
\newblock In \emph{Proceedings of the IEEE/CVF Conference on Computer Vision
  and Pattern Recognition}, 12113--12122.

\bibitem[{Yun et~al.(2019)Yun, Han, Oh, Chun, Choe, and Yoo}]{yun2019cutmix}
Yun, S.; Han, D.; Oh, S.~J.; Chun, S.; Choe, J.; and Yoo, Y. 2019.
\newblock Cutmix: Regularization strategy to train strong classifiers with
  localizable features.
\newblock In \emph{Proceedings of the IEEE International Conference on Computer
  Vision}, 6023--6032.

\bibitem[{Zhang et~al.(2017)Zhang, Cisse, Dauphin, and
  Lopez-Paz}]{zhang2017mixup}
Zhang, H.; Cisse, M.; Dauphin, Y.~N.; and Lopez-Paz, D. 2017.
\newblock mixup: Beyond empirical risk minimization.
\newblock \emph{arXiv preprint arXiv:1710.09412} .

\bibitem[{Zhong et~al.(2020)Zhong, Zheng, Kang, Li, and Yang}]{zhong2020random}
Zhong, Z.; Zheng, L.; Kang, G.; Li, S.; and Yang, Y. 2020.
\newblock Random Erasing Data Augmentation.
\newblock In \emph{AAAI}, 13001--13008.

\end{thebibliography}

\end{document}